\documentclass[pdflatex,sn-mathphys-num,iicol]{sn-jnl}

\usepackage{graphicx}%
\usepackage{multirow}%
\usepackage{multicol}%
\usepackage{makecell}%
\usepackage{amsmath,amssymb,amsfonts}%
\usepackage{amsthm}%
\usepackage{mathrsfs}%
\usepackage[title]{appendix}%
\usepackage{xcolor}%
\usepackage{textcomp}%
\usepackage{manyfoot}%
\usepackage{booktabs}%
\usepackage{algorithm}%
\usepackage{algorithmicx}%
\usepackage{algpseudocode}%
\usepackage{listings}%

\theoremstyle{thmstyleone}%
\newtheorem{theorem}{Theorem}
\newtheorem{proposition}[theorem]{Proposition}%

\theoremstyle{thmstyletwo}%
\newtheorem{assumption}{Assumption}%

\theoremstyle{thmstylethree}%

\begin{document}

\title[Article Title]{Beyond Dataset Distillation: Lossless Dataset Concentration via Diffusion-Assisted Distribution Alignment}


\author[1,2]{\fnm{Tongfei} \sur{Liu}}\email{liutongfei22@mails.ucas.ac.cn}
\equalcont{These authors contributed equally to this work.}

\author*[1,2]{\fnm{Yufan} \sur{Liu}}\email{yufan.liu@ia.ac.cn}
\equalcont{These authors contributed equally to this work.}

\author*[1,3]{\fnm{Bing} \sur{Li}}\email{bli@nlpr.ia.ac.cn}

\author[1,2,4]{\fnm{Weiming} \sur{Hu}}\email{wmhu@nlpr.ia.ac.cn}

\affil*[1]{\orgdiv{State Key Laboratory of Multimodal Artificial Intelligence Systems}, \orgname{Institute of Automation, Chines Academy of Sciences}, \orgaddress{\street{95 Zhongguancun East Road}, \city{Beijing}, \postcode{100190}, \state{Beijing}, \country{China}}}

\affil[2]{\orgdiv{School of Artificial Intelligence}, \orgname{University of Chinese Academy of Sciences}, \orgaddress{\street{1 Yanqihu East Road}, \city{Beijing}, \postcode{101408}, \state{Beijing}, \country{China}}}

\affil[3]{\orgname{PeopleAI, Inc.}, \orgaddress{\street{68 Zhichun Road}, \city{Beijing}, \postcode{100098}, \state{Beijing}, \country{China}}}

\affil[4]{\orgdiv{School of Information Science and Technology}, \orgname{ShanghaiTech University}, \orgaddress{\street{393 Huaxia Middle Road}, \city{Shanghai}, \postcode{201210}, \state{Shanghai}, \country{China}}}


\abstract{The high cost and accessibility problem associated with large datasets hinder the development of large-scale visual recognition systems. Dataset Distillation addresses these problems by synthesizing compact surrogate datasets for efficient training, storage, transfer, and privacy preservation. The existing state-of-the-art diffusion-based dataset distillation methods face three issues: lack of theoretical justification, poor efficiency in scaling to high data volumes, and failure in data-free scenarios. To address these issues, we establish a theoretical framework that justifies the use of diffusion models by proving the equivalence between dataset distillation and distribution matching, and reveals an inherent efficiency limit in the dataset distillation paradigm. We then propose a Dataset Concentration (DsCo) framework that uses a diffusion-based Noise-Optimization (NOpt) method to synthesize a small yet representative set of samples, and optionally augments the synthetic data via "Doping", which mixes selected samples from the original dataset with the synthetic samples to overcome the efficiency limit of dataset distillation. DsCo is applicable in both data-accessible and data-free scenarios, achieving SOTA performances for low data volumes, and it extends well to high data volumes, where it nearly reduces the dataset size by half with no performance degradation.}




\maketitle

\section{Introduction}\label{sec1}

The establishment of large-scale visual recognition systems requires a large amount of high-fidelity data. However, the excessively large datasets incur prohibitive costs in model training and data storage, and their accessibility is often limited by transfer overhead and the growing privacy and security regulations. To address these problems, Dataset Distillation (Wang, 1811~\cite{1811DD}) (DD, also known as Dataset Condensation, DC) has emerged as a promising solution towards secure and efficient data utilization. It aims to synthesize an extremely small surrogate dataset for the original large target dataset while preserving its essential information. This significantly reduces the resource demands associated with training, storage, and transfer. Moreover, it also enables a privacy-preserving data-sharing paradigm by releasing a synthetic dataset instead of the original target dataset that may contain sensitive information, circumventing the data accessibility problem.

Among existing dataset distillation methods, diffusion-model-based dataset distillation methods (Abbasi, 2024~\cite{D3M}; Su, 2024~\cite{D4M}; Du, 2023~\cite{Minimax}; Chen, 2025~\cite{IGD}) that synthesize samples with pre-trained diffusion models (Rombach, 2022~\cite{Rombach}) have demonstrated state-of-the-art (SOTA) training performances and remarkable distillation efficiency on large-scale, high-resolution datasets, such as the image classification benchmark, ImageNet-1k (Deng, 2009~\cite{imagenet}). For instance, the Minimax-IGD (Chen, 2025~\cite{IGD}) method has distilled the ImageNet-1k dataset into a synthetic dataset of 50 items per class (IPC), reducing about $95\%$ of the data with a performance degradation of no more than $10\%$ on the ResNet-18 (He, 2016~\cite{resnet}) model. 

Despite the great success of the dataset distillation methods in reducing costs associated with large datasets and improving data accessibility, there remain some limitations in these methods. Firstly, the current dataset distillation methods typically work under extremely small surrogate set sizes of no more than 100 items per class (IPC) on large-scale datasets such as the ImageNet-1k due to increasing synthesis costs. It is unclear whether the dataset distillation paradigm can be extended to high-IPC settings on large-scale, high-resolution datasets, raising concerns about its consistency and reliability across different data regimes. Secondly, most methods require full access to the target dataset, failing in data-free scenarios where the target data is inaccessible for privacy or safety reasons. Thirdly, while the current SOTA methods use a pre-trained diffusion model to synthesize samples, the motivation for the use of diffusion models is based on heuristics and not analytically justified, raising doubts regarding the reliability and trustworthiness of the methods.

This work addresses these limitations by establishing a theoretical framework to analyze the dataset distillation task. The framework demonstrates that the dataset distillation task is equivalent to a distribution matching problem under mild conditions. Since the diffusion model is trained for distribution alignment, the motivation for using diffusion models to synthesize surrogate datasets is justified. The theoretical analysis further reveals an inherent random sampling bias in diffusion-based sampling that causes a distribution misalignment that cannot be mitigated via guidance functions or fine-tuning. Moreover, an illustrative analysis under the proposed framework reveals a fundamental efficiency bottleneck in the dataset distillation paradigm that stems from the presence of "far-apart" samples in the target dataset, making it inefficient to extend the dataset distillation methods to high-IPC settings. 

In light of these theoretical insights, we propose a \textbf{D}ata\textbf{s}et \textbf{Co}ncentration (DsCo) framework. It synthesizes a small dataset via a diffusion-based Noise-Optimization (NOpt) method with mitigated random sampling bias, which is applicable in both data-accessible and data-free scenarios. To overcome the fundamental efficiency limit of the dataset distillation paradigm, DsCo optionally augments the synthetic samples with selected "far-apart" samples from the target dataset via a "Doping" procedure, which is controlled by a "Dope Trigger". The adaptability of the DsCo framework ensures robust performance across various data volumes and various accessibility conditions.

Experiments across multiple datasets demonstrate that DsCo achieves state-of-the-art performance across multiple datasets and settings, and a cost analysis is performed to demonstrate the superior efficiency of DsCo compared to existing open-source diffusion-based methods in both high-IPC and low-IPC regimes. Further, the extended high-IPC experiments demonstrate that the concentrated datasets achieve lossless dataset concentration performances even when reducing the dataset by half, demonstrating strong reliability for the large-scale compression. In data-free scenarios, it outperforms all existing data-free dataset distillation methods, offering a practical path to learning from inaccessible data while safeguarding privacy.

Beyond technical metrics, this work has a significant broader impact: it enables efficient model training with reduced energy consumption, facilitates privacy-compliant data sharing while keeping the sensitive original dataset inaccessible, and enables effective surrogate dataset synthesis in the absence of the original dataset, contributing to global data accessibility.

This work substantially extends our previous work: Noise-Optimized Distribution Distillation for Dataset Condensation (Liu, 2025~\cite{NODD}), which is published in Proceedings of the 33rd ACM International Conference on Multimedia. The previous work identified the random sampling bias and proposed the Noise-Optimization framework to mitigate it in the data-free scenario. The data-free Noise-Optimization method was termed the "NODD" method in our previous work. In this work, we make the following extensions in theory, methodology, and experiments:
\begin{itemize}
    \item We establish a new theoretical framework to analyze the task of dataset distillation, justifying the use of diffusion models by proving the equivalence of dataset distillation to distribution matching, identifying an inherent random sampling bias in diffusion synthesis, and proving a fundamental efficiency limit in dataset distillation. 
    \item We propose a unified DsCo framework that encloses the NODD proposed in our previous work (Liu, 2025~\cite{NODD}) along with the newly proposed NOpt, Doping, and Dope Trigger to handle data-accessible and data-free scenarios under both high-IPC and low-IPC regimes.
    \item We perform extensive evaluation and ablation experiments to demonstrate the high-IPC scalability, superior concentration performance, and compelling cost-efficiency of the DsCo framework.
\end{itemize}

\section{Related Work}
\subsection{Diffusion Model}
The diffusion model is a generative model that synthesizes samples by reversing a diffusion process, which degenerates the samples in a target dataset into random noise. The process is called "denoising". It was first proposed by Sohl-Dickstein (2015~\cite{sohl}) and improved by Ho (2020~\cite{Ho}) to generate samples of high quality. The modern latent diffusion models are proposed by Rombach (2022~\cite{Rombach}), which encodes the high-resolution samples into compressed latent codes with a pre-trained VAE (Kingma, 2013~\cite{VAE}), so that the diffusion model is trained in the latent space and synthesizes the small-scale latent codes instead of the high-resolution samples, significantly reducing its training and synthesis costs. 

To control the sample synthesis process, the classifier guidance (Dhariwal, 2021~\cite{dhariwal2021diffusion}) and classifier-free guidance (Ho, 2022~\cite{cfg}) methods have been proposed to steer the sample synthesis in the denoising process. These methods modify the transformation from the diffusion model outputs to the mean and standard deviations of the denoised samples at each denoising step by adding a guidance term to the transformation, thus modifying the step-wise denoised sample statistics to the desired directions. In particular, the classifier-free guidance method has been widely used in modern text-to-image tasks.

In this work, we perform a theoretical analysis to demonstrate that diffusion models naturally fulfill a dataset distillation objective that aims to synthesize informative surrogate samples for a target dataset, and an improved Noise-Optimization method is proposed to improve their performance on this task.

\subsection{Dataset Distillation}
The task of Dataset Distillation (DD) aims to synthesize a small surrogate dataset for the given target dataset, so that the downstream model trained on the surrogate dataset demonstrates as strong a performance as possible on the downstream task. Based on the main idea, the dataset distillation methods can be categorized into five broad categories: direct meta-learning methods, surrogate meta-learning methods, distribution-matching methods, patchwork methods, and generative methods.

\paragraph{Direct Meta-Learning}
In the early days, the dataset distillation task was formulated into a meta-learning problem, which optimizes the performance of downstream models trained on the synthetic dataset. A line of work uses the direct meta-learning techniques to resolve the meta-learning problem. These methods iteratively optimize the synthetic dataset. At each iteration, the synthetic dataset is used to train a specific downstream model for a few steps with the training trajectory preserved, and the trained downstream model is used to compute a meta-loss on the target dataset, which is the classification loss of the model. The gradient of the meta-loss is back-propagated through the training trajectory via the classical Backpropagation Through Time (BPTT) technique (Rumelhart, 1986~\cite{BPTT}). The original dataset distillation method (Wang, 1811~\cite{1811DD}) firstly introduced this paradigm upon the introduction of the dataset distillation task, but the training trajectories are confined to only a few steps, inducing poor distillation performances. The LinBa method (Deng, 2022\cite{2206LinBa}) further extended this method to long trajectories by reconstructing the training trajectories with stored model checkpoints. RaT-BPTT (Feng, 2023~\cite{2311RatBPTT}) proposed a Random Truncated Window BPTT method that improves the distillation performance while reducing the synthesis costs. This line of work has successfully synthesized informative datasets, but the prohibitive computational resource requirement for the repetitive unrolling of training trajectories limits their applicability in larger datasets.

\paragraph{Surrogate Meta-Learning}
A series of works have proposed a range of surrogate methods to substitute the computationally cumbersome meta-learning techniques. The trajectory-matching methods (Cazenavette, 2022~\cite{MTT}; Cui, 2023~\cite{TESLA}; Du, 2023~\cite{FTD}; Li, 2024~\cite{ATT}) propose a surrogate meta objective for the synthetic dataset, which aligns the training trajectories of the same model trained on the target dataset and the synthetic dataset being optimized. The gradient of the meta loss is then back-propagated to the synthetic dataset by BPTT, or through a linearized substitute trajectory (Li, 2024~\cite{ATT}). Another line of works termed gradient-matching methods (Zhao, 2021~\cite{GM}; Zhao, 2021~\cite{DSA}; Kim, 2022~\cite{IDC}; Liu, 2023~\cite{DREAM}) propose another surrogate meta objective, which aligns the training gradients of the model snapshots on its training trajectory on the target dataset with the training gradients on the synthetic dataset. These methods can be viewed as a one-step variant of the trajectory matching methods. Apart from these methods, the linear methods (Nguyen, 2020~\cite{KIP}; Loo, 2022~\cite{RFAD}; Zhou, 2022~\cite{FRePo}; Yu, 2024~\cite{Teddy}) substitute the lengthy training trajectories by linearized approximations, such as the kernel ridge regression (Nguyen, 2020~\cite{KIP}) and Taylor-expanded trajectories (Yu, 2024~\cite{Teddy}). These surrogate methods reduced the synthesis cost significantly compared to the direct meta-learning methods, but the meta-learning nature of these methods results in strong architecture overfitting, such that the distilled dataset demonstrates degraded performances on architectures that are unseen during the distillation process.

\paragraph{Distribution Matching}
Another line of work reformulates the dataset distillation task as a distribution matching task. These methods are termed distribution-matching methods (Zhao, 2023~\cite{DM}; Wang, 2022~\cite{CAFE}; Zhang, 2024~\cite{DANCE}; Yin, 2023~\cite{SRe2L}; Yin, 2023~\cite{CDA}). They align the feature distributions of the synthetic dataset to that of the target dataset, with various kinds of features used in the alignment, such as the NNGP-kernel associated random features (Zhao, 2023~\cite{DM}), the various features extracted by classifiers on their training trajectories (Wang, 2022~\cite{CAFE}; Zhang, 2024~\cite{DANCE}), and the Batch-Normalization statistics stored in pre-trained classifiers (Yin, 2023~\cite{SRe2L}; Yin, 2023~\cite{CDA}; Shao, 2024~\cite{G-VBSM}; Zhou, 2024~\cite{SC-DD}; Shao, 2024~\cite{EDC}; Shen, 2025~\cite{DELT}). These methods further reduce the synthesis costs as they do not require back-propagation through training trajectories, while achieving comparable performances to contemporary works. 

So far, all the methods discussed above adopt a pixel-level optimization paradigm, which initializes the synthetic dataset and optimizes it through an iterative optimization procedure. Since the pixel-level optimization makes no constraints on the joint distribution of pixels, the pixels tend to be independently optimized, resulting in high-frequency noises in the synthesized samples that impair the generalization performance of the synthesized samples.

\paragraph{Patchwork Methods}
Recently, a series of dataset distillation works have proposed to crop important patches from the target dataset and stitch those patches into new samples while discarding the unimportant patches left. RDED (Sun, 2024~\cite{RDED}) firstly proposed this cropping-and-stitching paradigm, and DDPS (Zhong, 2024~\cite{DDPS}) further developed this technique with the help of diffusion models to demonstrate improved performances. These methods are significantly faster than all previous methods because no optimization is required in them, and they demonstrate satisfactory distillation performances with the help of the relabeling technique. However, their performances are sub-optimal compared to a series of recent works discussed below.

\paragraph{Generative Methods}
A line of work has explored the possibility of utilizing generative models in the dataset distillation task. Some early works (Zhao, 2023~\cite{ITGAN}; Wang, 2023~\cite{DiM}; Huang, 2021~\cite{GDD}; Li, 2024~\cite{GlobalStructure}) replace the synthetic datasets with generative models, and some other methods (Cazenavette, 2023~\cite{GLaD}; Moser, 2024~\cite{LD3M}) optimize the latent codes of the generative models to synthesize informative samples. These methods adopt a "generate-optimize" paradigm where the whole synthetic dataset is repetitively generated and used in the computation of classical dataset distillation objectives of meta-learning or distribution-matching dataset distillation methods. The repetitive synthesis of the samples incurs a prohibitive synthesis cost in these methods, making them unsuitable for large-scale datasets. 

In contrast to previous generative methods, a recent line of works (Abbasi, 2024~\cite{D3M}; Su, 2024~\cite{D4M}; Yuan, 2023~\cite{TDSDM}; Du, 2023~\cite{Minimax}; Chen, 2025~\cite{IGD}) use the denoising process of a pre-trained diffusion model to synthesize informative synthetic samples. In these methods, the synthesized samples are only synthesized once in a single denoising process, significantly boosting the synthesis efficiency. Among these methods, the Minimax (Du, 2023~\cite{Minimax}) method fine-tunes a pre-trained diffusion model for improved distribution alignment and synthesizes samples with random noise. IGD (Chen, 2025~\cite{IGD}) proposes an influence guidance function that stems from the classical gradient-matching method, which steers the denoising process of diffusion models to synthesize training-effective samples. OT (Cui, 2025~\cite{OT}) method proposes to solve an optimal transport problem at each denoising step and incorporate the solution into the guidance function of the step, and design specific soft labels for the synthesized samples to boost the distillation performance. These methods have demonstrated state-of-the-art (SOTA) dataset distillation performances on high-resolution datasets such as the ImageNet-1k and its two subsets, ImageNette and ImageWoof (Jeremy, 2019~\cite{nette}), with satisfactory distillation costs.

\section{Theoretical Analysis}
This section establishes a theoretical framework to analyse the dataset distillation problems. Section~\ref{sec:RKHS} illustrates the general theoretical framework and the key assumption made in the analysis. Section~\ref{sec:DDmm} demonstrates that the dataset distillation task is equivalent to solving the distribution matching problem in this framework. Then, Section~\ref{sec:DDDiff} illustrates that the distribution matching objective can be naturally fulfilled by generative diffusion models (Rombach, 2022~\cite{Rombach}). Subsequently, Section~\ref{sec:randomsamplebias} demonstrates that there exists a random sampling bias in the DDPM denoising process commonly adopted by existing diffusion-based dataset distillation methods. Further, Section~\ref{sec:faa} uncovers an inherent limitation of the current dataset distillation paradigm that limits its efficiency in scaling to high surrogate data volumes.

\subsection{The Memorize-Generalize Picture}\label{sec:RKHS}
This work studies a general task of composing a small surrogate dataset for a given target dataset, so that a downstream model trained on the surrogate dataset performs as well as one trained on the target dataset when applied to the downstream task. Specifically, the theoretical analysis in this work involves three components: a target sample set of $N_\mathcal{T}$ samples, denoted as $\mathcal{T}\equiv \{x^{\tau}\}_{\tau=1}^{\tau=N_\mathcal{T}}, x^{\tau}\in\mathbb{R}^d$, with sample indices $\tau=1,\ldots,N_\mathcal{T}$; a surrogate sample set of $N_\mathcal{S}$ samples, denoted as $\mathcal{S}\equiv \{x^s\}_{s=1}^{N_\mathcal{S}}, x^s\in\mathbb{R}^d$, with sample indices $s=1,\ldots,N_\mathcal{S}$; a model $\Phi$ that memorizes $\mathcal{S}$ through training to make predictions for $\mathcal{T}$, which represents the neural network used in the downstream task. If $\Phi$ makes a correct prediction for $x^{\tau}$ after memorizing $\mathcal{S}$, we say that $\Phi$ recognizes $x^{\tau}$.

\paragraph{Feature Dissimilarity and Kernel Function}
Consider an arbitrary positive-definite symmetric kernel function $k(x^a,x^b)$ for two arbitrary samples $x^a, x^b \in \mathbb{R}^d$. According to the Moore-Aronszajn Theorem, the kernel function is associated with a projection $\psi$ that projects arbitrary samples $x^a,x^b$ into features $f^a,f^b=\psi(x^a), \psi(x^b)$ in a Reproducing Kernel Hilbert Space (RKHS) $\mathcal{H}$ of dimension $d'\rightarrow \infty$, which satisfies
\begin{equation}
    k(x^a, x^b) = \langle f^a,f^b\rangle_\mathcal{H}.
\end{equation}
That is, the inner product $\langle\cdot,\cdot\rangle_\mathcal{H}$ of the projected features in the RKHS equals the kernel function. Consequently, the L-2 distance between $f^a$ and $f^b$ can be formulated as \begin{equation}
    ||f^a - f^b||^2_\mathcal{H} = k(x^a,x^a) + k(x^b,x^b) - 2k(x^a,x^b).
\end{equation}
If the kernel function is shift-invariant (i.e., $k(x^a,x^b)$ is solely determined by $(x^a-x^b)$), $k(x,x)$ is a constant for any $x$. Therefore, the feature distance can be reformulated as \begin{equation}
    ||f^a - f^b||^2_\mathcal{H} = \mathrm{Constant} -2k(x^a,x^b).
\end{equation}
Hence, for two arbitrary samples $x^a, x^b$, their feature dissimilarity, as measured by the L-2 distance between their projected features $f^a,f^b$, monotonically decreases with the kernel function $k(x^a,x^b)$.

\paragraph{Chance of Recognition}
When the model $\Phi$ memorizes a surrogate sample $x^s$ through training, its generalization capability enables it to recognize a target sample $x^{\tau}$ that bears a certain degree of feature similarity to $x^s$. The greater the feature similarity between $x^{\tau}$ and $x^s$, the more likely $x^{\tau}$ is recognized by memorizing $x^s$. Therefore, denoting the chance of recognizing $x^{\tau}$ by memorizing $x^s$ as $p_\mathrm{rec}(x^{\tau}|x^s)$, it is reasonable to assume
\begin{align}
            &p_\mathrm{rec}(x^{\tau}|x^s) > p_\mathrm{rec}(x^{\tau'}|x^s) \nonumber\\ 
            &\iff k(x^s,x^{\tau}) > k(x^s,x^{\tau'}),
\end{align}
where $x^s\in \mathcal{S}$, and $x^{\tau},x^{\tau'}\in\mathcal{T}$. Equivalently, the corresponding L-2 distances of features in the RKHS, denoted as $d_2(f^s,f^{\tau})$ and $d_2(f^s,f^{\tau'})$ with $f^s, f^{\tau}, f^{\tau'} = \psi(x^s), \psi(x^{\tau}), \psi(x^{\tau'})$, satisfy 
\begin{align}
    &p_\mathrm{rec}(x^{\tau}|x^s) > p_\mathrm{rec}(x^{\tau'}|x^s) \nonumber \\
    &\iff d_2(f^s, f^{\tau}) < d_2(f^s, f^{\tau'}).
\end{align}

\paragraph{Summary and Illustration}

\begin{figure}
  \centering
  \includegraphics[width=1.0\linewidth]{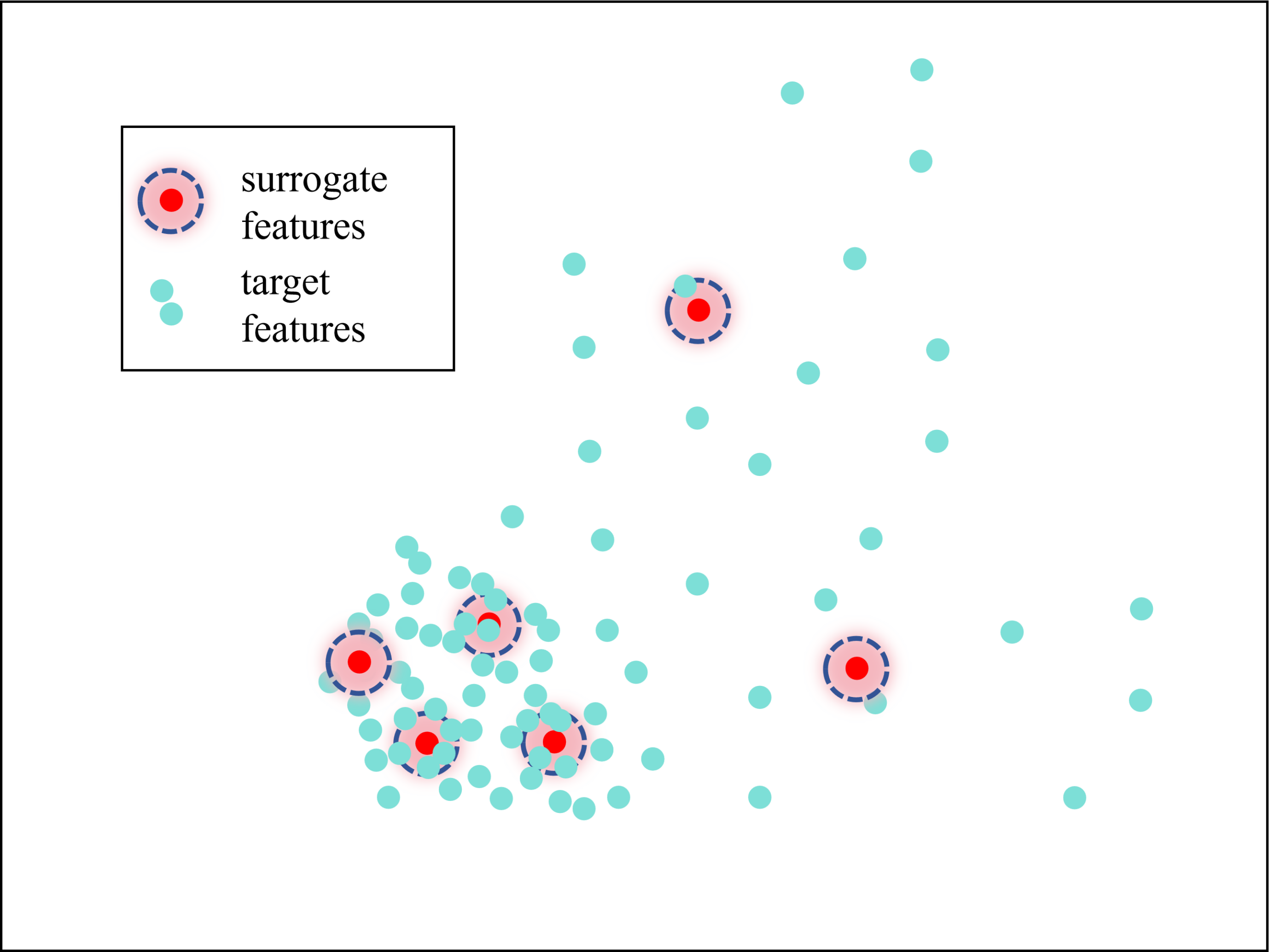}
  \caption{The graphical illustration of the target features and surrogate features plotted in the RKHS.}
  \label{fig:RKHS}
\end{figure}

In summary, the above construction can be reformulated into an assumption: 
\begin{assumption}\label{assum1}
For a model $\Phi$ which makes predictions on $\mathcal{T}$ by memorizing samples in $\mathcal{S}$, we assume that there exists a shift-invariant positive-definite kernel function $k(\cdot,\cdot)$, such that the chance of recognizing (i.e., making a correct prediction on) a sample $x^{\tau}\in\mathcal{T}$ by memorizing $x^s\in\mathcal{S}$ monotonically increases with increasing kernel function value $k(x^{\tau}, x^s)$.
\end{assumption}
Under this assumption, the Reproducing Kernel Hilbert Space (RKHS) associated with $k(\cdot,\cdot)$ can be described by the memorize-generalize picture, which is illustrated graphically by Figure~\ref{fig:RKHS}. In the figure, the projected features of the target set $\mathcal{T}$ and the surrogate set $\mathcal{S}$ in RKHS are scattered and denoted as the cyan and red dots, respectively. The fading pink region around each red dot denotes the chance of recognition associated with the corresponding surrogate sample, which fades away as the distance from the corresponding surrogate feature increases. The dashed circles indicate the maximal distance from the surrogate feature, beyond which the chance of recognizing a target sample by memorizing the corresponding surrogate sample is negligible. The picture provides an intuitive exposition of the relationship between the surrogate set, the target set, and the model, incentivizing the subsequent analysis.

\subsection{Dataset Distillation as a Distribution Matching Problem}\label{sec:DDmm}
As defined in the previous literature~\cite{1811DD}, the task of dataset distillation is to synthesize a small set of samples that can replace the target dataset to train a downstream model to achieve a high performance on the downstream task. In the Memorize-Generalize picture, this task can be reformulated into synthesizing a surrogate dataset $\mathcal{S}$ so that as many samples from $\mathcal{T}$ as possible can be recognized by memorizing $\mathcal{S}$. 

As previously demonstrated, the chance of recognizing a target sample $x^{\tau}$ by memorizing a surrogate sample $x^s$ monotonically decreases with the L-2 distance $d_2(f^s,f^{\tau})$. Considering the whole surrogate set, the chance of $x^{\tau}$ being recognized can be formulated as 
\begin{equation}
p_\mathrm{rec}(x^{\tau}|\mathcal{S}) = h\left(\sum_s^{N_\mathcal{S}}g(||f^{\tau} - f^s||^2)\right),
\end{equation}
where $g(\cdot):\mathbb{R}^+\rightarrow\mathbb{R}^+$ is an arbitrary monotonically decreasing function, $h(\cdot):\mathbb{R}^+\rightarrow \mathbb{R}^+$ is a normalizing function that normalizes the sum into the range $(0,1)$. Consider the continuous probability distribution of $\mathcal{T}$ and $\mathcal{S}$, denoted as $P_\mathcal{T}\equiv p_\mathcal{T}(x^{\tau})$ and  $P_\mathcal{S}\equiv p_\mathcal{S}(x^s)$, respectively. Their corresponding feature distributions in RKHS are expressed as $p_\mathcal{T}^f(f^{\tau})$ and $p^f_\mathcal{S}(f^s)$, with $p_\mathcal{T}^fdf^{\tau}=p_\mathcal{T}dx^{\tau}$ and $p_\mathcal{S}^fdf^s=p_\mathcal{S}dx^s$. Adopting this continuous expression and substituting the samples with the features in the RKHS, the expected chance of recognition 
in $\mathcal{T}$ is thus expressed as \begin{align}
    &\mathbb{E}_{x^{\tau}\sim P_\mathcal{T}}[p_\mathrm{rec}(x^{\tau}|\mathcal{S})] = \nonumber \\ &\int_{\mathbb{R}^{d'}}p_\mathcal{T}^f(f^{\tau})\cdot h\left(\int_{\mathbb{R}^{d'}}g(||f^{\tau}-f^s||)p^f_\mathcal{S}(f^s)df^s\right)df^{\tau}.
\end{align}
The task of dataset distillation is equivalent to maximizing the above expected chance with respect to $P_\mathcal{S}^\mathrm{RKHS}\equiv p_\mathcal{S}^f(f^s)$ under the constraint $\int p_\mathcal{S}^f(f^s)df^s=1$ and $p_\mathcal{S}^f(f^s)\geq0$. Denoting the above expected chance as $J(p_\mathcal{S}^f)$, The corresponding functional with Lagrangian multiplier $\lambda$ is thus 
\begin{equation}
    \mathcal{L}[p_\mathcal{S}^f] = J(p_\mathcal{S}^f) + \lambda(1-\int p_\mathcal{S}^f(f^s)df^s).
\end{equation}
Setting its variational derivative with respect to $p_\mathcal{S}^f$ to 0, the maximal chance of recognition is attained when the following condition is satisfied:
\begin{align}
    \int_{\mathbb{R}^{d'}}p_\mathcal{T}^f(f^{\tau}) h'\left((g\ast p_\mathcal{S}^f)(f^{\tau})\right)& g(||f^{\tau} - f^s||)df^{\tau} = \lambda, \nonumber \\
    & \forall f^s \in \mathrm{supp}(P_\mathcal{S}^\mathrm{RKHS}),
\end{align}
where $g\ast p_\mathcal{S}^f$ is the convolution between $g(\cdot)$ and $p_\mathcal{S}^f(\cdot)$. The normalizing function $h(\cdot)$ can be linear. In this case, the condition is simplified to 
\begin{equation}(g\ast p_\mathcal{T}^f)(f^s) = \mathrm{Constant}(\lambda)\quad \forall f^s\in \mathrm{supp}(P_\mathcal{S}^\mathrm{RKHS}).
\end{equation}
Further, as $g(||f^s-f^{\tau}||)$ is shift-invariant in RKHS, the above condition can be satisfied iff $p_\mathcal{S}^f(f) \propto p_\mathcal{T}^f(f), ~\forall f\in\mathbb{R}^{d'}$. For the two probability density functions, this indicates that 
\begin{equation}
    p_\mathcal{S}^f(f) = p_\mathcal{T}^f(f) \quad \forall f\in \mathbb{R}^{d'}.
\end{equation}
That is, the feature probability of $\mathcal{S}$ must be aligned with that of $\mathcal{T}$ to maximize the expected chance of recognition. In this case, $\mathbb{E}[f^s] = \mathbb{E}[f^{\tau}]$, thus the Maximal Mean Discrepancy associated with the kernel $k(\cdot,\cdot)$ between $\mathcal{S}$ and $\mathcal{T}$ is 0. This indicates that $P_\mathcal{S}=P_\mathcal{T}$ in the sample space $\mathbb{R}^d$.

To summarize, under Assumption~\ref{assum1}, the dataset distillation task is equivalent to solving a distribution matching problem. Therefore, the objective of dataset distillation can be reformulated into synthesizing a surrogate dataset to maximize the distribution alignment between the surrogate set and the target dataset. Formally, this conclusion can be formulated as follows:
\begin{proposition}\label{prop:1}
For a model $\Phi$ which makes predictions on $\mathcal{T}$ by memorizing samples in $\mathcal{S}$, if Assumption~\ref{assum1} holds, the maximization of the total expected chance of recognition of the samples in $\mathcal{T}$ is equivalent to the minimization of the continuous distribution discrepancy between $\mathcal{S}$ and $\mathcal{T}$.
\end{proposition}
In Appendix~\ref{app:extendDM}, this proposition is extended to apply under a broader condition:
\begin{proposition}\label{prop:2}
    For a model $\Phi$ which makes predictions on $\mathcal{T}$ by memorizing samples in $\mathcal{S}$, if there exists an invertible transformation that maps the samples into a feature space, in which there exist a shift-invariant positive-definite kernel function whose value monotonically increases with the chance of recognizing a target sample $x^{\tau}\in\mathcal{T}$ by memorizing a surrogate sample $x^s\in\mathcal{S}$, the maximization of the total expected chance of recognition of the samples in $\mathcal{T}$ is equivalent to the minimization of the continuous distribution discrepancy between $\mathcal{S}$ and $\mathcal{T}$.
\end{proposition}
This is equivalent to stating that the dataset distillation task is essentially resolving a distribution matching problem.
\subsection{Generative Diffusion Model for Distribution Matching}\label{sec:DDDiff}
Consider an arbitrary sample set $\mathcal{X}\equiv\{x^n\}_{n=1}^{N}$ of $N$ samples. A pre-trained autoencoder encodes the samples to latent codes $\mathcal{Z}\equiv\{z^n\}_{n=1}^N$ of reduced dimension, $z^n\in \mathbb{R}^{C,H,W}$, which can be decoded back to $\mathcal{X}$ via the decoder. The encoded latent codes can be diffused via an iterative diffusion process, so that at step $t$ of the diffusion process, the diffused latent code $z_t$ for an arbitrary latent code $z_0$ is
\begin{equation}\label{eqn:diffuse}
    z_t = \sqrt{\alpha_t}z_0 + \sqrt{1-\alpha_t}\epsilon,\quad\epsilon\sim N(0,I),
\end{equation}
where $\alpha_t$s are the scheduling coefficients for different $t$. At the last step $T$ of diffusion, the diffused latent code follows the Gaussian distribution, $z_T \sim N(0,I)$.

A latent diffusion model denoted as $\Psi$ learns to reverse the diffusion process, so that an arbitrary noise $z_T\sim N(0,I)$ can be mapped to a denoised latent code $z_0$ via a series of denoising steps, $t=T,\ldots,0$. At a single denoising step $t$, given the denoised sample $z_{t+1}$ from the previous step, the diffusion model $\Psi$ predicts the mean $\mu_t$ and standard deviation $\sigma_t$ of the denoised sample $z_t$ via a non-linear transformation, formulated as
\begin{equation}\label{eqn:trans}
    (\mu_t, \sigma_t) = \mathrm{trans}(\Psi(z_{t+1},t)),
\end{equation}
where $\mathrm{trans}$ maps the output of the diffusion model to the statistics of $z_t$. After that, a random noise $\epsilon_t\sim N(0,I)$ is sampled from the Gaussian distribution to predict $z_t$ as follows:
\begin{equation}\label{eqn:denoise}
    z_t = \mu_t + \sigma_t \cdot \epsilon_t.
\end{equation}

To analyze the diffusion model in the memorize-generalize picture, consider a latent diffusion model $\Psi_\mathcal{T}$ trained on $\mathcal{Z}^\mathcal{T}\equiv\{z^{\tau}\}_{\tau=1}^{N_\mathcal{T}}$. Given a set of $M$ noise tensors, $\mathcal{Z}_T \equiv \{z_T^m\}_{m=1}^M,~z_T^m \sim N(0,I)$, the denoising process of $\Psi$ consecutively maps the set of tensors into a set of denoised latent codes, $\mathcal{Z}_0\equiv\{z_0^m\}_{m=1}^M$. In the continuous limit, the probability distributions of the two sets $\mathcal{Z}^\mathcal{T}$ and $\mathcal{Z}_0$ can be formulated as $p_{\mathcal{Z}^\mathcal{T}}(z^{\tau})$ and $p_{\mathcal{Z}_0}(z_0^m)$. As demonstrated by the previous literature~\cite{Ho}, $\Psi_\mathcal{T}$ minimizes the discrepancy between the two distributions by minimizing the distribution discrepancy between the denoised and diffused samples at each denoising and diffusion step $t$. Therefore, the distribution discrepancy between the samples $\mathcal{T}$ and $\mathcal{X}_0$ decoded from the latent codes $\mathcal{Z}, \mathcal{Z}_0$ is minimized for a fixed decoder. 

In summary, the decoded denoised sample set $\mathcal{X}_0$ of a latent diffusion model $\Psi_\mathcal{T}$ trained on the latent code set $\mathcal{Z}^\mathcal{T}$ encoded from $\mathcal{T}$ has a probability distribution with minimized discrepancy from that of $\mathcal{X}$. Hence, $\mathcal{X}_0$ naturally fulfills the equivalent dataset distillation objective in the memorize-generalize picture, serving as an effective surrogate set $\mathcal{S}$.

\subsection{The Random Sampling Bias}\label{sec:randomsamplebias}
The previous analysis theoretically demonstrates that a diffusion model $\Psi_\mathcal{T}$ trained on the target dataset $\mathcal{T}$ naturally generates effective surrogate samples for $\mathcal{T}$, fulfilling the dataset distillation objective. However, in practice, when the size of the surrogate set is limited, the alignment is sub-optimal, resulting in reduced training performance. 

To analyze the misalignment, consider a particular step $t$ in the denoising process that generates a latent surrogate set $Z^\mathcal{S}_0\equiv \{z_0^s\}_{s=1}^{N_\mathcal{S}}$ of size $N_\mathcal{S}$. At step $t$, a latent code denoted as $z_t^{\tau}$ is diffused from its corresponding target latent code, $z^{\tau}\in Z^\mathcal{T}$, using Equation~\ref{eqn:diffuse}, and a denoised code $z_t^s$ is computed with Equation~\ref{eqn:denoise}:
\begin{equation}
    z_t^s = \mu_t^s + \sigma_t^s\cdot \epsilon_t^s.
\end{equation}
Therefore, the probability distribution of $z_t^s$ in the continuous limit, denoted as $p(z_t^s)$, satisfies the following relationship:
\begin{align}
    p(z_t^s)dz_t^s &= p(\mu_t^s, \sigma_t^s, \epsilon_t^s)d\mu_t^sd\sigma_t^sd\epsilon_t^s \nonumber \\
    &=p(\mu_t^s, \sigma_t^s)p(\epsilon_t^s)d\mu_t^sd\sigma_t^sd\epsilon_t^s
\end{align}
Hence, $p(z_t^s)$ is independently determined by the joint distribution of $(\mu_t^s, \sigma_t^s)$ and the distribution of $\epsilon_t^s$. Ideally, $\epsilon_t^s$ strictly follows the Gaussian distribution $\mathcal{N}(0,I)$, and $(\mu_t^s, \sigma_t^s)$ follow an ideal distribution denoted as $p^*(\mu_t, \sigma_t)$, so that $p(z_t^s)$ strictly aligns to the diffused latent code probability $p(z_t^{\tau})$. However, in practice, both distributions can be compromised. The statistics distribution $p(\mu_t^s, \sigma_t^s)$ may deviate from the ideal $p^*(\mu_t, \sigma_t)$ due to systematic bias, and the noise distribution may deviate from $\mathcal{N}(0,I)$ due to random errors.

In light of Equation~\ref{eqn:trans}, the systematic bias in $p(\mu_t^s, \sigma_t^s)$ stems from three sources: the bias in the denoised sample distribution from the previous step $p(z_{t+1}^{\tau})$, the inherent bias of the diffusion model $\Psi_\mathcal{T}$, and the bias in the mapping function $\mathrm{trans}(\cdot)$. Previous works on diffusion-based dataset distillation have attempted to mitigate the bias in $\Psi_\mathcal{T}$ by a Minimax fine-tuning (Gu, 2024~\cite{Minimax}) and calibrate $\mathrm{trans}(\cdot)$ by modifying its guidance function (Chen, 2025~\cite{IGD}; Chen, 2025~\cite{OT}). However, even if the systematic bias in $p(\mu_t^s, \sigma_t^s)$ has been mitigated, the random sampling bias in $\epsilon_t^s$ persists due to the random noise sampling, as illustrated in Appendix~\ref{app:noiseMonteCarlo} through a simple Monte-Carlo experiment. Therefore, the denoised latent code distribution $p(z_t^s)$ still differs from the target diffused latent code probability $p(z_t^{\tau})$ regardless of the calibration proposed by previous works. In the subsequent denoising step $t-1$, the bias in $p(z_t^s)$ results in a distribution bias in $(\mu_{t-1}, \sigma_{t-1})$, which colludes with the random sampling bias in $\epsilon_{t-1}$ to produce even more biased denoised latent code distribution $p(z_{t-1}^s)$. The accumulation of the distribution bias results in a biased sampling of $\mathcal{Z}^\mathcal{S}_0$, resulting in performance degradations of the decoded surrogate dataset $\mathcal{S}$.

In summary, the random noise sampling at each step of the denoising process of a diffusion model results in a bias in the step-wise denoised latent code distribution that accumulates across the denoising process, resulting in impaired training performance of the synthesized surrogate dataset. In order to mitigate this problem, this work proposes a Noise-Optimization framework to manually align the denoised latent code distribution with the diffused target probability distribution, as detailed in Section~\ref{sec:NOpt}.

\subsection{The Far-Apart Samples}\label{sec:faa}
In the memorize-generalize picture, the chance of recognition monotonically decreases with the L-2 distance between the projected features in RKHS. In this case, we can define a specific recognition threshold $r_\mathrm{rec}$ associated with the downstream model in the RKHS, as illustrated by the dashed rings in Figure~\ref{fig:RKHS}. For a surrogate feature $f^s$, if a target feature $f^{\tau}$ lies further away from it than $r_\mathrm{rec}$, the corresponding target sample $x^{\tau}$ is very unlikely to be recognized by memorizing $x^s$. The recognition threshold is a direct indicator of the generalization capability of the model, as models with larger $r_\mathrm{rec}$ could recognize more target samples by memorizing fewer surrogate samples.

For a model with recognition threshold $r_\mathrm{rec}$, consider a special kind of target samples whose feature distances from their nearest neighbors in the target set in RKHS are greater than $2\times r_\mathrm{rec}$. In this work, these samples are termed "far-apart samples". A key property that distinguishes these samples from other ordinary target samples is that, for an arbitrary surrogate sample, there can be at most one far-apart sample that is recognized by memorizing the surrogate sample. This is because the recognition threshold around an arbitrary surrogate feature in the RKHS can accommodate at most one far-apart sample. 

The existence of far-apart samples results in a natural limitation in the dataset distillation paradigm, which synthesizes surrogate samples to maximize the expected chance of recognition of the target dataset. That is, when all the non-far-away samples have been recognized by memorizing the surrogate dataset, the number of extra target samples recognized by synthesizing one extra surrogate sample cannot exceed one. In this case, since the cost of synthesizing one surrogate sample is significantly greater than directly sampling a target sample, it is no longer efficient to synthesize more surrogate samples. Instead, the far-away samples should be directly sampled to complement the synthesized samples. In this work, we term this improved strategy beyond dataset distillation as "Doping".

\section{Method}
\begin{figure*}
  \centering
  \includegraphics[width=1.0\linewidth]{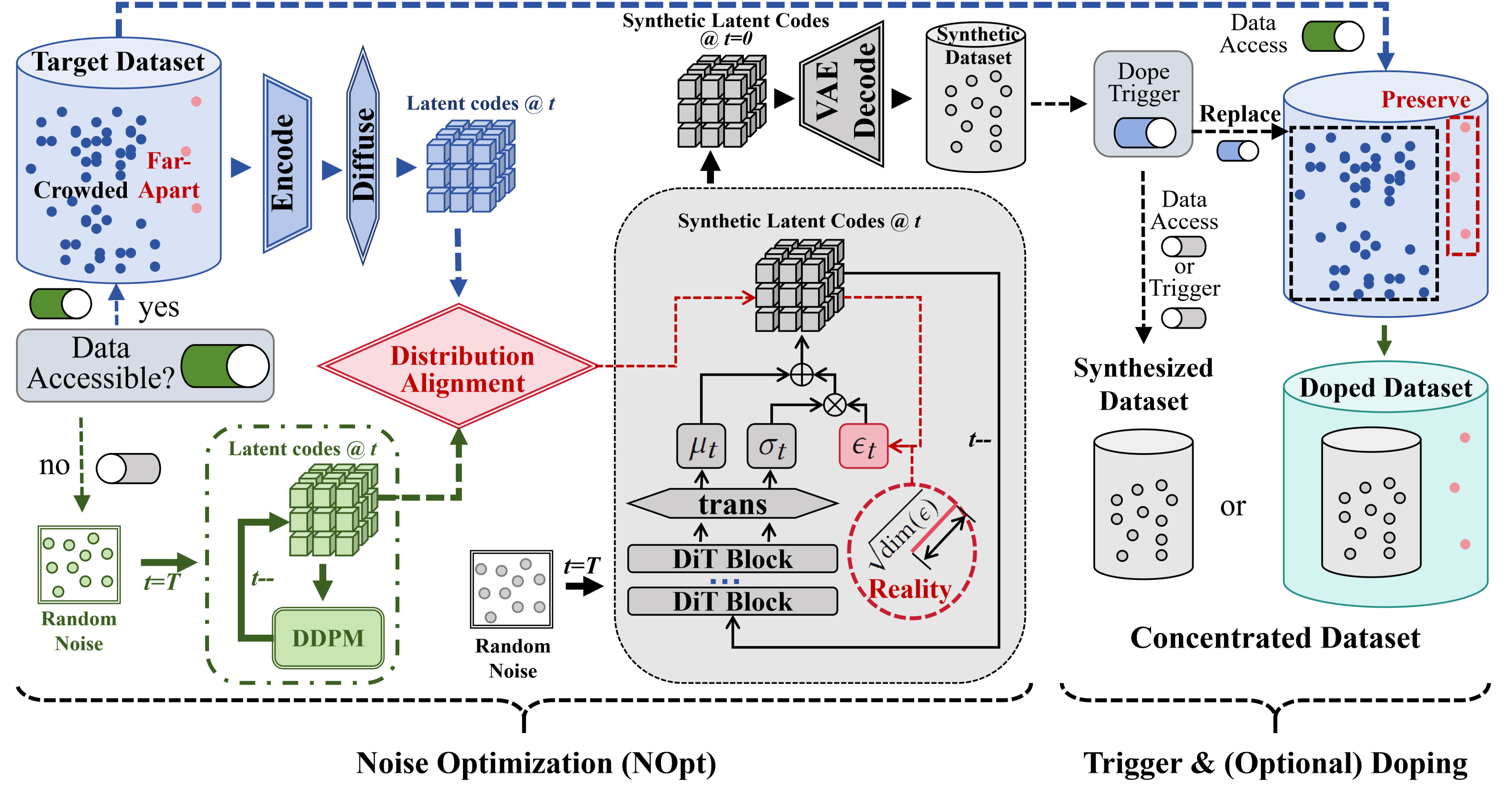}
  \caption{The Dataset Concentration framework. Enclosed in the gray region is the iterative denoising process of NOpt; the dashed red arrows indicate the gradient flow during each noise-optimization. }
  \label{fig:DsCo}
\end{figure*}
As indicated by the theoretical analysis, the state-of-the-art diffusion-based dataset distillation paradigm has two fundamental limitations: the persistent random sampling bias present in denoising synthesis and the inefficiency in synthesizing surrogate samples for far-apart samples. Moreover, in practice, the target dataset can be inaccessible due to various reasons, such as safety, privacy, copyright, and transfer costs. The inaccessibility renders the majority of existing dataset distillation methods inapplicable, and the applicable ones ($\mathrm{SRe^2L}$ and vanilla DiT) suffer from sub-optimal training performances.

In this work, we propose a \textbf{D}ata\textbf{s}et \textbf{Co}ncentration (DsCo) framework to resolve these problems. The DsCo framework is illustrated in Figure~\ref{fig:DsCo}. As illustrated in the figure, for a target dataset, DsCo firstly synthesizes a compact and informative sample set using an innovative diffusion-based \textbf{N}oise-\textbf{Opt}imization (NOpt) method that mitigates the sampling bias and enforces a step-wise distribution alignment by optimizing the random noise tensors upon generating synthetic samples through the denoising process of the diffusion model. In particular, the distribution alignment can be enforced in both data-accessible and data-free scenarios, where the target distribution information is obtained either by encoding and diffusing the accessible target samples or by generating step-wise latent codes using the denoising process of the diffusion model pre-trained on the inaccessible target dataset. After the denoising process, the denoised synthetic latent codes are decoded into synthetic samples with a decoder. Subsequently, if the samples have reached the efficiency limit of the synthesis paradigm (indicated by the "Dope Trigger" in the figure), the far-apart samples in the target dataset are identified and selected to complement the synthesized sample set. This process is termed Doping. It effectively replaces the crowded samples (denoted by the blue dots in the figure) that can be represented by the synthetic samples with the set of synthetic samples (denoted by the gray cylinder), and preserves the far-apart samples denoted by the pink dots. The Doping process enables the extension of the surrogate dataset to high IPCs that are beyond the extensibility of the current dataset distillation methods. In the end, the DsCo framework composes a surrogate dataset for the target dataset under both data-accessible and data-free scenarios. If Doping is triggered and the target data can be accessed, the concentrated dataset is a mixture of selected far-apart target samples and the synthetic samples generated with data-accessible distribution alignment. Otherwise, the concentrated dataset is the set of synthetic samples generated by NOpt with data-accessible or data-free distribution alignment.

\subsection{The Noise-Optimization Framework}\label{sec:NOpt}
In light of the theoretical analysis in Section~\ref{sec:randomsamplebias}, a generative diffusion model is trained so that its denoising process generates samples whose distribution in the continuous limit aligns with the target sample distribution. Hence, a diffusion model can be used to generate a set of surrogate samples to replace the target sample set to train downstream models while preserving the model performance. However, there exist multiple biases in the denoising process, including the systematic biases in the diffusion model and the denoising transformation, and the random bias incurred by random noise sampling at each denoising step. The biases impair the distribution alignment, resulting in sub-optimal training performance of the surrogate set. Among the biases, the random bias stemming from random noise sampling cannot be mitigated by modifying the diffusion model or the denoising transformation. Therefore, this work proposes the \textbf{N}oise-\textbf{Opt}imization (NOpt) method that enforces distribution alignment by optimizing the noise tensors in the denoising process. 

The left of Figure~\ref{fig:DsCo} illustrates the proposed Noise-Optimization (NOpt) method. It is an improved DDPM denoising process that synthesizes the set of synthesized latent codes $\mathcal{Z}^\mathcal{S}\equiv\{z^s\}_{s=1}^{N_\mathcal{S}}$ through iterative denoising steps with optimized noise tensors, where $s$ is the sample index.

Specifically, the gray region enclosed by the dashed black lines illustrates the denoising steps of NOpt in detail. At each step $t$, the diffusion model predicts the set of sample-wise means and standard deviations of the denoised samples, $\{(\mu_t^s,\sigma_t^s)\}_{s=1}^{N_\mathcal{S}}$ from the denoised samples from the previous step, denoted as $\mathcal{Z}_{t+1}^\mathcal{S}$. Meanwhile, a set of Gaussian noise tensors denoted as $\mathcal{E}_t^\mathcal{S}\equiv\{\epsilon_t^s\}_{s=1}^{N_\mathcal{S}}$ is sampled randomly. After that, the noise tensors are optimized to enforce feature distribution alignment under a reality constraint. At each step of the optimization of the noise tensors, the set of denoised samples, $\mathcal{Z}_t^\mathcal{S}$, are computed with Equation~\ref{eqn:trans} from $\mathcal{E}_t^\mathcal{S}$ and $\{(\mu_t^s,\sigma_t^s)\}_{s=1}^{N_\mathcal{S}}$, and projected into features with a random projector. The set of features is denoted as $\mathcal{F}_t^\mathcal{S}$. Subsequently, it enforces the distribution alignment constraint under either data-accessible or data-free scenario, as detailed in Section~\ref{sec:da_align} and~\ref{sec:df_align}. Meanwhile, a reality constraint applies a geometric regularization on the noise tensors, $\mathcal{E}_t^\mathcal{S}$. The noise tensors are then optimized to minimize the feature distribution dissimilarity and the reality constraint. After the noise optimization, the final denoised latent codes computed from the optimized noise tensors are passed into the next denoising step $t-1$. 

\subsubsection{Reality Constraint}\label{sec:reality}
The Noise-Optimization framework optimizes the set of noise tensors $\mathcal{E}_t^\mathcal{S}$ to minimize the feature distribution dissimilarity between the denoised and diffused samples. However, as the optimization proceeds, the noise tensors may deviate significantly from the Gaussian distribution, resulting in out-of-distribution denoised samples. To avoid this issue, we propose a reality constraint that regularizes the noise to follow the Gaussian distribution.

For illustration, consider an arbitrary $d$-dimensional noise sample $ u= (u_1,\ldots, u_d)^T\in \mathbb{R}^d$ from the $d$-dimensional Gaussian distribution. Each element, $u_i$, independently follows the standard normal distribution, $u_i \sim \mathcal{N}(0,1)$. The L-2 norm of $u$, denoted as $\mathrm{norm}(u)$, can be reformulated as \begin{equation}\mathrm{norm}(u) = \sqrt{\sum_i^d (u_i-0)^2} = \sqrt{d \cdot \mathrm{Var}(\hat{u})},\end{equation} where $\mathrm{Var}(\hat{u})$ is the sample variance of $d$ samples $\{\hat{u}\}$ independently sampled from $\mathcal{N}(0,1)$. As $d$ increases, the law of large numbers ensures that $\mathrm{Var}(\hat{u})\rightarrow 1$, hence $\mathrm{norm}(u) \rightarrow \sqrt{d}$. Therefore, the norm of a sample from the high-dimensional standard normal distribution is almost surely close to $\sqrt{d}$. Since the high-dimensional standard normal distribution is spherically homogeneous, we argue that any $d$-dimensional tensor $u$ with $\mathrm{norm}(u)\approx\sqrt{d}$ is very likely to be sampled from the $d$-dimensional normal distribution. 

The above analysis indicates that we can regularize any noise tensor $\epsilon_t\in\mathbb{R}^d$ to follow the $d$-dimensional Gaussian distribution by minimizing the discrepancy between its norm and $\sqrt{d}$. For fast convergence, we adopt the "absolute and square" function, defined as \begin{equation}
    \mathrm{absn}^2(x)\equiv |x|+x^2,~ \forall x\in \mathbb{R},
\end{equation} which provides non-diminishing gradient at all $x$ except for $x=0$ (where its gradient is set to 0). 

In summary, for the set of noise tensors, $\mathcal{E}_t^\mathcal{S}=\{\epsilon_t^s\},~\epsilon_t^s\in\mathbb{R}^{C,H,W}$, the reality constraint is \begin{equation}\label{eqn:reality}
\mathcal{L}^t_{\mathrm{real}} \equiv \sum_{s=1}^{N_\mathcal{S}}\mathrm{absn^2}(\mathrm{norm}(\epsilon_t^s)-\sqrt{CHW}).\end{equation}.

\subsubsection{Random Feature Projection}\label{sec:projector}
As proposed by ~\cite{CNN}, a key characteristic of visual features is the spatial shift-invariance. That is, the spatial shift of a visual feature contributes little to the identification of the feature. Latent DD~\cite{latentdd} has demonstrated that this property is also present in the encoded latent space of VAE. Therefore, the latent denoised samples also possess this property of shift-invariance. Consequently, the variation in the exact location of a visual feature in a denoised latent code $z_t$ contributes little to the information capacity of the latent code. When the number of samples in the condensed set is limited, the variations in the spatial locations of the visual features in the target dataset impose an extraneous spatial distribution constraint that detrimentally interferes with the information capacity of the synthesized samples. 

To mitigate this problem, this work proposes to project each latent code, $z_t\in \mathbb{R}^{C,H,W}$, into a feature of significantly reduced spatial dimensions $(K\ll H,L\ll W)$ and increased channel dimension $J\gg C$, denoted as $f_t \in \mathbb{R}^{J,K,L}$. The projection is conducted with a randomly initialized 3-layer Convolutional Neural Network (CNN) architecture similar to the encoder architecture used in the small-scale VAE (Kingma, 2013~\cite{VAE}), as detailed in Appendix~\ref{app:projector}. In particular, to save computational cost in feature projection, we adopt the grouped convolutional layers (Krizhevsky, 2012~\cite{alexnet}) in the projector, making it equivalent to multiple random projectors that independently and simultaneously project the latent codes into a concatenated set of independent random features.

The benefit of projecting the latent codes into features using a randomly initialized convolutional neural network is beyond the mitigation of spatial redundancy. In fact, it also facilitates the subsequent distribution alignment by decoupling the entangled latent code distribution into a series of channel-wise feature distributions that are nearly independent. As demonstrated by the previous literature (Novak, 2018~\cite{NNGP}), for a single layer of a convolutional neural network with $m$ channels, the correlation between different channels of its output diminishes at the speed of $O(1/\sqrt{m})$ as $m$ increases. Since the projector used in this work is wide, with $J\gg C$, the correlation between different channels of $f_t$ is strongly suppressed. Therefore, the projected feature $f_t$ can be viewed as $J$ independent features concatenated together.

\subsubsection{Data-Accessible Distribution Alignment}~\label{sec:da_align}
As illustrated in Section~\ref{sec:randomsamplebias}, the random sampling bias results in an accumulating distribution mismatch between the denoised latent codes and the diffused latent codes at each step. Therefore, the distribution of the denoised latent codes at each step is biased. To mitigate this problem, this work proposes to manually align the denoised and diffused latent code distributions at each step $t$ by minimizing a distribution alignment loss $\mathcal{L}_\mathrm{align}^t$, which enforces the distribution alignment between the two sets of projected features of the denoised and diffused latent codes.

\paragraph{Diffused Latent Code Distribution}
The diffused target latent codes are obtained by diffusion. For each latent code $z^\tau$ with index $\tau$ in the set of target latent codes $\mathcal{Z}^\mathcal{T}$, we compute its corresponding diffused latent code with Equation~\ref{eqn:diffuse} with random noise for at least five times, and we group all the diffused target latent codes to form a set, denoted as $\mathcal{Z}_t^\mathrm{Diff}\equiv \{z_t^{\tau'}\}_{\tau'=1}^{N_\mathrm{Diff}}$. $N_\mathrm{Diff}$ denotes the total number of diffused latent codes in $\mathcal{Z}_t^\mathrm{Diff}$, which is automatically adjusted to the minimal value that satisfies $N_\mathrm{Diff}\bmod N_\mathcal{S}=0$ and $N_\mathrm{diff}/N_\mathcal{S} \geq 5$ to facilitate the subsequent distribution alignment. $\tau'$ denotes the index of the diffused latent code.

\paragraph{Channel-Wise Distribution Alignment}
As previously argued in Section~\ref{sec:projector}, the two sets of latent codes are projected into feature tensors, $\mathcal{F}_t^\mathcal{S}\equiv \{f_t^s\}_{s=1}^{s=N_\mathcal{S}}$ and $\mathcal{F}_t^\mathrm{Diff}\equiv \{f_t^{\tau'}\}_{\tau'=1}^{\tau'=N_\mathrm{Diff}}$, with $f_t^s, f_t^{\tau'} \in \mathbb{R}^{J,K,L}$. The correlation between the channels of the features is strongly suppressed by the random projector. Therefore, the feature distribution alignment can be enforced in a channel-wise manner, where each channel of the two sets of features is independently aligned. 

For simplicity, we further reduce the spatial dimensions of the features to $(1,1)$ with a Global Average Pooling (Lin, 2013~\cite{NiN}) layer, so that each channel has only one dimension. Further, the two sets of pooled features are normalized by the element-wise mean and standard deviation of the pooled features of $\mathcal{Z}_t^\mathrm{Diff}$ for scale-invariance. The pooled and normalized features are denoted as $\hat{\mathcal{F}}_t^\mathcal{S}\equiv \{\hat{f}_t^s\}_{s=1}^{N_\mathcal{S}}$ and $\hat{\mathcal{F}}_t^\mathrm{Diff}\equiv \{\hat{f}_t^{\tau'}\}_{\tau'=1}^{N_\mathrm{Diff}}$, where $\hat{f}_t^{s}, \hat{f}_t^{\tau'}\in\mathbb{R}^{J}$. The two sets of $j$-th elements of the normalized features are denoted as $\hat{\mathcal{F}}_t^{\mathcal{S},j}\equiv \{\hat{f}_t^{s,j}\}_{s=1}^{N_\mathcal{S}}$ and $\hat{\mathcal{F}}_t^{\mathrm{Diff},j}\equiv \{\hat{f}_t^{\tau',j}\}_{\tau'=1}^{N_\mathrm{Diff}}$.

For each channel $j$, the alignment is enforced by minimizing the discrepancy between each $\hat{f}_t^{s,j}$ and its corresponding interpolated percentile in $\hat{\mathcal{F}}_t^{\mathrm{Diff},j}$, which is the solution to the one-dimensional optimal transport problem (Villani, 2021~\cite{OptimalTransport}). In practice, $\hat{\mathcal{F}}_t^{\mathrm{Diff},j}$ is sorted in ascending order and subsequently chunked into $N_\mathcal{S}$ groups of equal size, $N_\mathrm{chunk} = N_\mathrm{Diff}/N_\mathcal{S}$, which is bound to be an integer as $N_\mathrm{Diff}\bmod N_\mathcal{S}=0$. Subsequently, each percentile corresponding to the respective $\hat{f}_t^{s,j}$ is computed by taking the average of $\hat{f}_t^{\tau',j}$s in the corresponding group. Finally, the channel-wise distribution alignment for channel $j$ is enforced by minimizing the following loss:
\begin{equation}
    \mathcal{L}_{\mathrm{ch},j}^t = N_\mathcal{S}\sum_{s=1}^{N_\mathcal{S}} \mathrm{absn}^2(\hat{f}_t^{s,j} - \frac{\sum_{\tau'=(s-1)\times N_\mathrm{chunk}}^{s \times N_\mathrm{chunk}}\hat{f}_t^{\tau',j}}{N_\mathrm{chunk}}),
\end{equation}
where the extra $N_\mathcal{S}$ corrects for the variation in the loss value with $N_\mathcal{S}$.
\paragraph{Alignment Enforcement}
In summary, the feature distribution alignment is enforced by minimizing the following distribution alignment loss: \begin{equation}\label{eqn:daalign}
    \mathcal{L}_\mathrm{align}^t \equiv \lambda_\mathrm{align}\sum_j^J\mathcal{L}_{\mathrm{ch},j}^t.
\end{equation}
The $\lambda_\mathrm{align}$ is a hyperparameter determining the strength of feature distribution alignment relative to the reality constraint.

\subsubsection{The Data-free Distribution Alignment}\label{sec:df_align}
In the data-free scenario, a data-free version of the distribution alignment is enforced without access to the target dataset. Since the samples in the target dataset cannot be accessed, the diffused latent codes at $t$ cannot be properly acquired. To resolve this problem, we approximate the feature distribution of the diffused target latent codes using the statistics of the features of the denoised latent codes, and apply data-free distribution alignment by maximizing the spatial occupation of the synthesized sample features under regularization of the approximated feature distribution.

\paragraph{Distribution Approximation} In order to approximate the diffused distribution, we generate a random template set of latent codes via the DDPM denoising process of the pre-trained diffusion model. The set of template denoised latent codes at step $t$ is denoted as $\mathcal{Z}_t^\mathrm{temp} \equiv \{z_t^i\}_{i=1}^{N_\mathrm{temp}}$, where $N_\mathrm{temp}$ is the number of template samples.

Subsequently, we project the template denoised latent codes $\mathcal{Z}_t^\mathrm{temp}$ into features, $\mathcal{F}_t^\mathrm{temp} \equiv \{f_t^i\}_{i=1}^{N_\mathrm{temp}},~f_t^i \in \mathbb{R}^{J,K,L}$, with the previously mentioned random projector, and calculate the first two statistics (i.e., the mean and standard deviation) of the set of projected template features. In particular, to remove the extraneous spatial information, the statistics are computed in a channel-wise manner. For an arbitrary set of $N$ features $\mathcal{F}\equiv \{f^i\}_{i=1}^{N}$ with shape $f\in\mathbb{R}^{J,K,L}$, element indices $j,k,l$, and feature index $i$, the channel-wise mean ($\mathrm{mean}^\mathrm{ch}(\mathcal{F})\in\mathbb{R}^{J}$) and standard deviation ($\mathrm{std}^\mathrm{ch}(\mathcal{F})\in\mathbb{R}^{J}$) are computed as following:
\begin{align}
&\mathrm{mean}^\mathrm{ch}(\mathcal{F})_j \equiv \frac{1}{NKL}\sum_{i,k,l}^{N_,K,L} f^i_{j,k,l}, \\
&\mathrm{std}^\mathrm{ch}(\mathcal{F})_j \equiv \sqrt{\frac{1}{NKL}\sum_{i,k,l}^{N,K,L}[f^i_{j,k,l} - \mathrm{mean}^\mathrm{ch}(\mathcal{F})_j]^2 }.
\end{align}
Further, for two arbitrary sets of features, $\mathcal{F}\equiv \{f^i\}_{i=1}^{N},~f^i \in \mathbb{R}^{J,K,L}$ and $\mathcal{F'}\equiv \{f^{m'}\}_{m'=1}^{N'},~f^{m'}\in \mathbb{R}^{J,K,L}$, we define a channel-wise cross-normalization function that normalizes a sample $f^i$ with element indices $j,k,l$ in $\mathcal{F}$ by the channel-wise statistics of $\mathcal{F}'$,  
\begin{equation}
    \mathrm{normalize}(f^i, \mathcal{F}')_{j,k,l} \equiv \frac{f^i_{j,k,l} - \mathrm{mean}^\mathrm{ch}(\mathcal{F'})_j}{\mathrm{std}^\mathrm{ch}(\mathcal{F}')_j}.
\end{equation}
Thereafter, we cross-normalize the projected features of the surrogate latent codes, $\mathcal{F}_t^\mathcal{S}$, with the projected template features, $\mathcal{F}_t^\mathrm{temp}$, to compute the data-free distribution alignment loss. The set of cross-normalized projected latent features is denoted as $\overline{\mathcal{F}}_t^\mathcal{S}\equiv\{\overline{f}_t^s\}_{s=1}^{N_\mathcal{S}}$. In practice, the distribution alignment is enforced by a maximal-occupation loss under the constraint of a channel-wise feature statistics regularization. 

\paragraph{Maximal Occupation}
The maximal occupation loss maximizes the spatial occupation of each cross-normalized projected feature, $\overline{f}_t^s$, in the projected feature space $\mathbb{R}^{J,K,L}$. The motivation is twofold. Firstly, maximizing the spatial occupation of each projected feature mitigates the sample overlapping problem illustrated in Section~\ref{sec:randomsamplebias}. Secondly, as demonstrated in the previous literature (Gu, 2024~\cite{Minimax}), a generative diffusion model trained on the target dataset tends to generate representative samples that reside in the dense regions of the target sample distribution, leading to insufficient diversity. Maximizing the spatial occupation of each latent code feature improves the diversity by increasing the feature dissimilarity among the samples.

In practice, the Maximal Occupation loss maximizes the cosine distance $d_\mathrm{cos}$ of each cross-normalized surrogate feature $\overline{f}_t^s$ from its nearest neighbor in $\overline{\mathcal{F}}_t^\mathcal{S}$, where $d_\mathrm{cos}(u,v)\equiv-(u\cdot v/|u||v|)$ for any $u,v\in\mathbb{R}^d$. For a generic sample $u$ of arbitrary shape from a set $\mathcal{U}$, its nearest neighbor $\mathrm{nn}(u)$ is
\begin{equation} 
\mathrm{nn}(u) \equiv \mathrm{arg~min}_{u'\in\mathcal{U}, u'\neq u}D_\mathrm{cos}(u, u'). \end{equation}
Therefore, the maximal-occupation loss, denoted as $\mathcal{L}_\mathrm{maxoc}^t$, is \begin{equation}
    \mathcal{L}_\mathrm{maxoc}^{t} \equiv -\sum_s^{N_\mathcal{S}}d_\mathrm{cos}(\overline{f}_t^s, \mathrm{nn}(\overline{f}_t^s)).
\end{equation}

\paragraph{Feature Statistics Regularization}
While the maximal occupation loss improves the feature diversity in the surrogate sample, excessive diversity results in a distribution mismatch between the surrogate set and the target dataset, as demonstrated in the previous literature (Gu, 2024~\cite{Minimax}). To avoid this problem, we enforce a distribution regularization on the statistics of the surrogate features, explicitly aligning the channel-wise mean and standard deviation of the set of surrogate features to those of the template features, which serve as approximations to the feature statistics of the diffused latent codes from the target dataset. Since the surrogate features have been cross-normalized by the statistics of the template features, the regularization is enforced by minimizing the discrepancy of the channel-wise mean and standard deviation of the set of cross-normalized features from 0 and 1, respectively. Therefore, the feature statistics regularization term of the data-free distribution alignment loss is 
\begin{equation}
\begin{aligned}
    \mathcal{L}^{t}_\mathrm{stats} \equiv& \sum_{j}^{J}\mathrm{absn}^2\left(\mathrm{mean}^\mathrm{ch}(\overline{\mathcal{F}}_t^\mathcal{S})_j-0\right)\\&+\sum_{j}^{J}\mathrm{absn}^2\left(\mathrm{std}^\mathrm{ch}(\overline{F}_t^\mathcal{S})_j-1\right).
\end{aligned}
\end{equation}

\paragraph{Alignment Enforcement}
In summary, the data-free distribution alignment loss can be formulated as \begin{equation}\label{eqn:dfalign}
    \mathcal{L}_\mathrm{align}^t \equiv \lambda_\mathrm{stats}\mathcal{L}_\mathrm{stats}^t + \lambda_\mathrm{maxoc}\mathcal{L}_\mathrm{maxoc}^t,
\end{equation}
where $\lambda_\mathrm{stats}$ and $\lambda_\mathrm{maxoc}$ are the respective hyperparameters regularizing the relative strengths of the two terms. In particular, the $\lambda_\mathrm{maxoc}$ is fixed to the maximal value that enables the synthesis of visually sensible samples when $\lambda_\mathrm{stats}$ is 0. For all experiments presented in this work, this value is 10.0.

\subsubsection{The Noise-Optimization Objective}
In light of the above formulation, at each denoising step $t$, the Noise-Optimization method optimizes the noise tensors to minimize a combined noise-optimization loss $\mathcal{L}_\mathrm{NOpt}$, which is defined as 
\begin{equation}\label{eqn:nopt}
    \mathcal{L}_\mathrm{NOpt}^t \equiv \mathcal{L}_\mathrm{real}^t + \mathcal{L}_\mathrm{align}^t ,
\end{equation}
where $\mathcal{L}_\mathrm{real}^t$ is defined in Equation~\ref{eqn:reality}, and the data-accessible and data-free versions of $\mathcal{L}_\mathrm{align}^t$ are defined in Equation~\ref{eqn:daalign} and Equation~\ref{eqn:dfalign}, respectively. 

\subsection{Dope Trigger and Doping}
In light of the analysis in section~\ref{sec:faa}, the far-apart samples are special samples whose nearest neighbors in the target dataset are beyond the generalization capability of the classifier. For these samples, it is uneconomic to synthesize representative samples, because one synthesized sample can cover at most one far-apart sample, yet synthesizing a surrogate sample incurs extra costs compared to directly sampling the exact target sample. Instead, this work proposes to identify and sample these samples to mix them up with the previously synthesized representative samples, forming a mixed dataset. However, there are two problems to be resolved: When to switch from synthesis to sampling? How to sample the far-apart samples? 

\subsubsection{The Dope Trigger}
The Dope Trigger describes a criterion on the number of samples synthesized, which determines when the switching from synthesis to sampling occurs. The criterion is deduced based on the reasoning below.

In light of the theoretical analysis, the distilled synthetic samples are synthesized to maximize the number of samples recognized by the model by minimizing their distribution discrepancy from the target dataset. In the RKHS picture, as illustrated in Figure~\ref{fig:RKHS}, the synthetic samples need to enclose as many target samples as possible into their recognition thresholds to increase the number of target samples recognized. Therefore, these synthetic samples tend to concentrate around the high-density regions of the target distribution. As the number of synthetic samples increases, the new synthetic samples tend to occupy the previously unoccupied regions with the highest target sample density. Therefore, the number of additional target samples recognized by the addition of synthetic samples monotonically decreases as the number of synthetic samples increases. Denoting a synthetic dataset of $M$ samples as $\mathcal{S}_M$, and the corresponding number of target samples recognized as $N_M$, the marginal gain of increasing the number of synthetic samples from $M_1$ to $M_2$ can be defined as \begin{equation}
    \frac{\Delta N}{\Delta M}\equiv\frac{N_{M_2}-N_{M_1}}{M_2-M_1},
\end{equation}
which monotonically decreases as $M_1$ increases. When $\frac{\Delta N}{\Delta M} > 1$,  the synthetic dataset has not fully covered the crowded samples, thus the Dope Trigger is off. When $\frac{\Delta N}{\Delta M} \leq 1$, adding one synthetic sample results in the recognition of at most one target sample, thus all the crowded samples have been recognized, and the unrecognized target samples are all far-apart samples. 

In summary, the Dope Trigger is essentially the condition that $\frac{\Delta N}{\Delta M} \leq 1$. When the Dope Trigger is on, the Doping method is adopted to incorporate the far-apart samples into the surrogate dataset.

\subsubsection{Doping}
Given a designated size of surrogate set measured in the number of items per class (IPC), the Dope Trigger determines whether the synthetic dataset has covered the crowded regions in the target sample distribution, and the Doping method identifies the far-apart samples to be selected. Since the chance of a target sample being recognized by a model diminishes as its RKHS feature distance from its nearest surrogate neighbor increases, the more confused the model is about the target sample, the further it is from the crowded region of the target sample, the more likely it is a far-apart sample. Hence, we can identify the far-apart samples by assessing the degree to which the model trained on the synthesized representative samples is confused about them, which is measured in the confusion score. 

The confusion score is determined as follows. Firstly, a downstream model, $\phi_\mathcal{S}$ is trained with the previously synthesized small dataset $\mathcal{S}$ and the corresponding labels provided by a model $\phi_\mathcal{T}$ pre-trained on the target dataset. Subsequently, a confusion score is computed for each sample in the target dataset based on the predictions made by both models. The confusion score measures the degree to which the decision made by $\phi_\mathcal{S}$ differs from $\phi_\mathcal{T}$. For a training sample $x_i \in \mathcal{T}$, the models trained on the target dataset and the synthesized dataset make post-softmax predictions, $p_\mathcal{T}(x_i) = (p_\mathcal{T}^1(x_i),\ldots,p_\mathcal{T}^C(x_i))^T$ and $p_\mathcal{S}(x_i) =(p_\mathcal{S}^1(x_i),\ldots,p_\mathcal{S}^C(x_i))^T$, respectively. The prediction made by $\phi_\mathcal{T}$ is $c_\mathcal{T}\equiv \mathrm{arg max}_c~p_\mathcal{T}^c$. The confusion score measures the degree to which $\phi_\mathcal{S}$ disagrees with this prediction: 
\begin{equation}
    \mathrm{Conf} \equiv p_\mathcal{S}^{c_\mathcal{T}} - \mathrm{max}_{c\neq c_\mathcal{T}}~p_\mathcal{S}^c.
\end{equation}

If a sample has a prediction made by $\phi_\mathcal{S}$ that disagrees with that made by $\phi_\mathcal{T}$, it is bound to have positive confusion scores, and those with unanimous hard predictions have negative confusion scores. This gives the samples with erroneous predictions a higher priority compared to those that have been correctly classified. In practice, the samples with the highest confusion scores are selected to complement the synthetic sample set.

\section{Experiments}
\subsection{Experimental Setup}
\subsubsection{Datasets}
We follow previous diffusion-based dataset condensation methods to evaluate the proposed framework on the ImageNet-1K (Deng, 2009~\cite{imagenet}) dataset and its two representative subsets, ImageNette and ImageWoof, proposed in (Jeremy, 2019~\cite{nette}). The ImageNet-1K dataset is a large-scale image classification dataset with 1000 classes of images, with at most 1300 training images and 50 test images in each class. The ImageNette and ImageWoof datasets are two representative subsets of ImageNet-1K. ImageNette is a simple subset that consists of 10 very dissimilar classes, while ImageWoof consists of ten dog breeds bearing significant visual similarity, making it difficult for classification tasks. All images are resized to $224\times224$ upon training classifiers.
\subsubsection{Evaluation Metrics}
This work follows previous literature in diffusion-based DD to evaluate the condensed or concentrated datasets with the top-1 classification accuracies of a variety of model architectures trained on the corresponding datasets. The architectures include ConvNet-6 (LeCun, 2002~\cite{CNN}), ResNet-AP10 (He, 2016~\cite{resnet}), ResNet-18, ResNet-34, ResNet-101, EfficientNet-B0 (Tan, 2019~\cite{Efficient}),  MobileNet-V2 (Sandler, 2018~\cite{mobilenet}), and DenseNet-121 (Huang, 2017~\cite{densenet}). Unless otherwise specified, the models are trained on the concentrated dataset five times, and the average evaluation performances are reported with the corresponding standard deviations.
\subsubsection{Implementation Details}
For the Dataset Concentration framework under both data-accessible and data-free scenarios, we follow previous diffusion-based DD methods to use a publicly available latent DiT (William, 2022~\cite{DiT}) model pre-trained on ImageNet-1K as the backbone diffusion model, which is available for download in the code repository associated with the DiT paper; the images are encoded and decoded with a pre-trained VAE model from Stable Diffusion (Robin, 2021\cite{sd}). For the evaluation of the composed or distilled surrogate sets, the training settings of the models are the same as in (Chen, 2025~\cite{IGD}), where all the models are trained with the AdamW (Loshchilov, 2017~\cite{AdamW}) optimizer, unless otherwise specified. The detailed implementations of the training of the models are specified in Appendix~\ref{app:implementation}. The use of the relabeling technique that provides informative soft labels for the distilled or composed samples using a pre-trained ResNet-18 model, which is commonly adopted in the Dataset Distillation literature (Yin, 2023~\cite{SRe2L}; Yin, 2023~\cite{CDA}; Sun, 2024~\cite{RDED}; Chen, 2025~\cite{IGD}), is specified in the corresponding analyses where applicable. The specific implementations of NOpt, Dope Trigger, and Doping are detailed below. 

\paragraph{NOpt Implementation Details} In the Noise-Optimization (NOpt) stage of datasets concentration, the surrogate noises, $\epsilon_t^s$s, are optimized for 200 steps by an SGD optimizer with a learning rate of 0.1 and a momentum of 0.9. The encoder-like feature projector is detailed in Appendix~\ref{app:projector}. For the data-free distribution alignment, 200 random denoised template samples are used to compute the feature statistics. The adjustable hyperparameters for the data-accessible and data-free distribution alignment, $\lambda_\mathrm{align}$ and $\lambda_\mathrm{stats}$, are set to 0.0005 and 0.001 for all the experiments in this work except in the ablation studies, where they are tuned to analyze the hyperparameter sensitivity.

\paragraph{Dope Trigger and Doping}
In this work, the Dope Trigger is determined by assessing the marginal gain of increasing the number of synthesized samples, $\frac{\Delta N}{\Delta M}\equiv\frac{N_{M_2}-N_{M_1}}{M_2-M_1}$. The synthetic dataset size $M$ is incremented in a schedule of (10IPC, 50IPC, 100IPC, 150IPC), and the corresponding marginal gains are computed to determine the Dope Trigger. For the ImageWoof and IMageNette, $\frac{\Delta N}{\Delta M}>0$ for 10IPC, 50IPC, and 100IPC, hence the Doping is triggered when the designated surrogate size is greater than 100IPC. For ImageNet-1k, Doping is triggered for surrogate sizes greater than 50IPC. The Doping in this work is performed by sampling the samples with the highest confusion scores, which are computed with the respective evaluation models, unless otherwise specified in the corresponding experiment.

\subsubsection{Baselines}
We compare the DsCo framework with a variety of existing dataset distillation methods, including the traditional DD methods (DM(Zhao, 2023~\cite{DM}), IDC(Kim, 2022~\cite{IDC}), $\mathrm{SRe^2L}$(Yin, 2023~\cite{SRe2L}), CDA (Yin, 2023~\cite{CDA}), EDC (Shao, 2024~\cite{EDC}), G-VBSM (Shao, 2024~\cite{G-VBSM}), SC-DD~\cite{SC-DD} (Zhou, 2024~\cite{SC-DD}), CV-DD (Cui, 2025~\cite{CVDD}), patchwork methods (RDED (Sun, 2024~\cite{RDED}), DDPS (zhong, 2024~\cite{DDPS})), and the diffusion-based generative DD methods ($\mathrm{D^3M}$Abbasi, 2024~\cite{D3M},$\mathrm{D^4M}$ (Su, 2024~\cite{D4M}), TDSDM (Yuan, 2023~\cite{TDSDM}), DiT (William, 2022~\cite{DiT}), Minimax (Gu, 2024~\cite{Minimax}), IGD (Chen, 2025~\cite{IGD}), and OT (Chen, 2025~\cite{OT})). In particular, for fair comparison, we reproduce the samples for DiT using the publicly available DiT checkpoint and report the reproduced evaluation performances. We label the reproduced DiT baseline with a $\dagger$ sign for identification. Besides, it is worth noting that EDC and OT use specifically tailored soft labels instead of the commonly used ResNet-18 soft labels for the ImageNet-1k experiments; thus, the two methods are not presented as baselines in the corresponding tables to avoid misleading comparisons.

\subsection{Performance Analysis}
\begin{table*}
\small
\setlength{\tabcolsep}{1.5pt}
\caption{Performance comparison on ImageNette under low-IPC setting. Best performances are denoted in \textbf{bold}.}
\label{tab:nette_lowipc}
\begin{tabular}{c|ccc|ccc|ccc}
\toprule
\multicolumn{10}{c}{ImageNette}\\
\hline
Model & \multicolumn{3}{c}{ConvNet6} & \multicolumn{3}{c}{ResNetAP-10} &  \multicolumn{3}{c}{ResNet-18} \\\hline
Full & \multicolumn{3}{c}{94.3$\pm$0.5} & \multicolumn{3}{c}{94.6$\pm$0.5} & \multicolumn{3}{c}{95.3$\pm$0.6} \\ \hline
IPC &10&50&100&10&50&100&10&50&100\\
\hline  \multicolumn{10}{c}{Data-Accessible}\\
\hline
Random & 46.0$\pm$0.5 & 71.8$\pm$1.2 & 79.9$\pm$0.8 &54.2$\pm$1.2&77.3$\pm$1.0&81.1$\pm$0.6&55.8$\pm$1.0&75.8$\pm$1.1&82.0$\pm$0.4 \\
DM & 49.8$\pm$1.1&70.3$\pm$0.8&78.5$\pm$0.8&60.2$\pm$0.7&76.7$\pm$1.1&80.9$\pm$0.7&60.9$\pm$0.7&75.0$\pm$1.0&81.5$\pm$0.4 \\
IDC-I&48.2$\pm$1.2&72.4$\pm$0.7&80.6$\pm$1.1&60.4$\pm$0.6&77.4$\pm$0.7&81.5$\pm$1.2&61.0$\pm$1.8&77.5$\pm$1.0&81.7$\pm$0.8\\
Minimax&58.2$\pm$0.9&76.9$\pm$0.8&81.1$\pm$0.3&63.2$\pm$1.0&78.2$\pm$0.7&81.5$\pm$1.0&64.9$\pm$0.6&78.1$\pm$0.6&81.3$\pm$0.7\\
DiT-IGD&61.9$\pm$1.9 & 80.9$\pm$0.9&84.5$\pm$0.7&66.5$\pm$1.1&81.0$\pm$1.2&85.2$\pm$0.8&67.7$\pm$0.3&80.4$\pm$0.8&84.4$\pm$0.8\\
Minimax-IGD&58.8$\pm$1.0&82.3$\pm$0.8&86.3$\pm$0.8&63.5$\pm$1.1&82.3$\pm$1.1&86.1$\pm$0.9&66.2$\pm$1.2&82.0$\pm$0.3&86.0$\pm$0.6\\
OT&\textbf{67.0$\pm$0.9}&83.1$\pm$1.0&86.5$\pm$0.5&68.0$\pm$0.3&83.8$\pm$0.6&86.4$\pm$0.6&69.1$\pm$1.9&\textbf{84.6$\pm$0.4}&85.9$\pm$0.2 \\
DsCo(ours)&65.7$\pm$0.8&\textbf{83.2$\pm$0.6}&\textbf{86.9$\pm$0.2}&\textbf{69.8$\pm$1.1}&\textbf{84.0$\pm$0.7}&\textbf{86.9$\pm$0.8}&\textbf{70.3$\pm$0.1}&83.4$\pm$0.9&\textbf{86.5$\pm$0.9}\\
\hline  \multicolumn{10}{c}{Data-Free}\\ \hline
$\mathrm{DiT}^\dagger$&61.5$\pm$0.9&74.3$\pm$0.9& 79.1$\pm$ 1.1 &64.8$\pm$0.3&75.5$\pm$0.6& 80.5$\pm$0.7 &64.2$\pm$0.9&74.9$\pm$0.1& 78.2$\pm$0.7\\
DFDsCo(ours)&63.0$\pm$1.2&80.3$\pm$0.2& 82.2$\pm$0.4 &67.0$\pm$1.7&81.8$\pm$0.9& 81.6$\pm$1.1 & 67.5$\pm$0.8&81.0$\pm$0.9& 79.0$\pm$1.0\\
\bottomrule
\end{tabular}
\end{table*}

\begin{table*}
\small
\setlength{\tabcolsep}{1.5pt}
\caption{Performance comparison on ImageWoof under low-IPC setting.Best performances are denoted in \textbf{bold}.}
\label{tab:woof_lowipc}
\begin{tabular}{c|ccc|ccc|ccc}
\toprule
\multicolumn{10}{c}{ImageWoof}\\
\hline
Model & \multicolumn{3}{c}{ConvNet6} & \multicolumn{3}{c}{ResNetAP-10} &  \multicolumn{3}{c}{ResNet-18} \\\hline
Full & \multicolumn{3}{c}{85.9$\pm$0.4} & \multicolumn{3}{c}{87.2$\pm$0.6} & \multicolumn{3}{c}{89.0$\pm$0.6} \\ \hline
IPC &10&50&100&10&50&100&10&50&100\\
\hline  \multicolumn{10}{c}{Data-Accessible}\\
\hline
Random & 25.2$\pm$1.1& 41.9$\pm$1.4 & 52.3$\pm$1.5 &31.6$\pm$0.8&50.1$\pm$1.6&59.2$\pm$0.9&30.9$\pm$1.3&54.0$\pm$0.8&63.6$\pm$0.5 \\
DM&27.6$\pm$1.2&43.8$\pm$0.8&50.1$\pm$0.9&29.8$\pm$1.0&47.8$\pm$1.2&59.8$\pm$1.3&30.2$\pm$0.6&53.9$\pm$0.7&64.9$\pm$0.7 \\
IDC-I&34.1$\pm$0.8&42.6$\pm$0.9&51.0$\pm$1.1&38.5$\pm$0.7&49.3$\pm$0.9&56.4$\pm$0.5&36.7$\pm$0.8&54.5$\pm$1.0&57.7$\pm$0.8\\
Minimax&33.5$\pm$1.4&50.7$\pm$1.8&57.1$\pm$1.9&39.6$\pm$1.2&59.8$\pm$0.8&66.8$\pm$1.2&42.2$\pm$1.2&60.5$\pm$0.5&67.4$\pm$0.7\\
DiT-IGD&35.0$\pm$0.8 & 54.2$\pm$0.7&61.1$\pm$1.0&41.0$\pm$0.8&62.7$\pm$1.2&69.7$\pm$0.9&44.8$\pm$0.8&62.0$\pm$1.1&70.6$\pm$1.8\\
Minimax-IGD&36.2$\pm$1.6&\textbf{55.7$\pm$0.8}&\textbf{63.0$\pm$1.8}&43.3$\pm$0.3&65.0$\pm$0.8&\textbf{71.5$\pm$0.8}&\textbf{47.2$\pm$1.6}&65.4$\pm$1.8&72.1$\pm$0.9\\
DsCo(ours)&\textbf{37.9$\pm$0.5}&53.7$\pm$0.7&\textbf{63.0$\pm$1.5}&\textbf{47.1$\pm$0.3}&\textbf{65.3$\pm$0.7}&71.2$\pm$0.3&45.2$\pm$1.0&\textbf{65.5$\pm$0.2}&\textbf{72.7$\pm$1.0}\\
\hline  \multicolumn{10}{c}{Data-Free}\\ \hline
$\mathrm{DiT}^{\dagger}$&33.3$\pm$0.3&47.9$\pm$1.0& 54.5$\pm$1.0&39.7$\pm$0.9&57.1$\pm$0.6& 63.1$\pm$1.4 &41.1$\pm$1.1&57.5$\pm$1.7& 62.2$\pm$ 0.9\\
DFDsCo(ours)&34.9$\pm$1.3&51.5$\pm$0.9& 57.6$\pm$0.6 &41.0$\pm$2.2&59.8$\pm$0.7& 66.6$\pm$1.1 &43.3$\pm$ 2.1& 59.9$\pm$1.2 & 61.2$\pm$1.0 \\
\bottomrule
\end{tabular}
\end{table*}

\begin{table*}
\setlength{\tabcolsep}{9pt}
\caption{Validation Accuracy vs IPC for Dataset Concentration on ImageNette.}
\label{tab:nette_highipc}
\begin{tabular}{c|cccccc|c}
\toprule
Dataset&\multicolumn{7}{c}{ImageNette}\\
\hline
IPC& 200& 300& 400& 700& 1000& 1300&\multirow{2}{*}{Full}\\
Ratio\%& 15.5& 23.3 & 31.0 & 54.3 & 77.6 & 100.8&\\
\hline Model&\multicolumn{6}{c}{ConvNet6}\\
\hline
Random& 92.6$\pm$0.8&92.6$\pm$1.1& 93.1$\pm$0.7& 94.7$\pm$0.4& 95.9$\pm$0.2&96.5$\pm$0.2&\multirow{2}{*}{96.5$\pm$0.2} \\
DsCo&94.0$\pm$0.7& 94.2$\pm$0.5& 94.5$\pm$0.6&95.7$\pm$0.5&96.5$\pm$0.2& 96.6$\pm$0.2&\\
\hline Model&\multicolumn{6}{c}{ResNetAP-10}\\
\hline
Random&92.7$\pm$1.1&94.4$\pm$0.8&94.5$\pm$0.9&96.5$\pm$0.4& 96.5$\pm$0.3&96.8$\pm$0.3&\multirow{2}{*}{96.8$\pm$0.3}\\
DsCo&93.5$\pm$0.5&94.0$\pm$0.6&95.4$\pm$0.5&97.2$\pm$0.3&97.1$\pm$0.1&97.1$\pm$0.3\\
\hline Model&\multicolumn{6}{c}{ResNet-18}\\
\hline
Random& 95.1$\pm$0.7 &95.6$\pm$0.9 & 96.1$\pm$0.9 & 97.0$\pm$0.4 & 97.1$\pm$0.3& 97.8$\pm$0.1&\multirow{2}{*}{97.8$\pm$0.1}\\
DsCo& 96.0$\pm$0.5 &96.1$\pm$0.4 & 96.9$\pm$0.5 & 97.9$\pm$0.1 & 97.8$\pm$0.3 & 98.0$\pm$0.2 &\\
\bottomrule
\end{tabular}
\end{table*}

\begin{table*}
\setlength{\tabcolsep}{9pt}
\caption{Validation Accuracy vs IPC for Dataset Concentration on ImageWoof.}
\label{tab:woof_highipc}
\begin{tabular}{c|cccccc|c}
\toprule
Dataset&\multicolumn{7}{c}{ImageWoof}\\
\hline
IPC& 200& 300& 400& 700& 1000& 1300&\multirow{2}{*}{Full}\\
Ratio\%& 16.0& 24.1 & 32.1 & 56.2 & 80.3 & 104.4&\\
\hline Model&\multicolumn{6}{c}{ConvNet6}\\
\hline
Random& 80.2$\pm$1.0& 80.1$\pm$1.0& 83.1$\pm$0.5& 87.4$\pm$0.6& 87.4$\pm$0.1& 88.3$\pm$0.3&\multirow{2}{*}{88.3$\pm$0.3}\\
DsCo& 81.0$\pm$0.5&82.7$\pm$0.5& 84.0$\pm$0.7&88.4$\pm$0.4 & 88.3$\pm$0.4&88.3$\pm$0.2& \\
\hline Model&\multicolumn{6}{c}{ResNetAP-10}\\
\hline
Random& 83.1$\pm$0.6& 83.1$\pm$0.3& 86.0$\pm$0.2& 88.4$\pm$0.5& 88.0$\pm$0.2&90.2$\pm$0.3 &\multirow{2}{*}{90.2$\pm$0.3}\\
DsCo& 85.9$\pm$0.6&85.8$\pm$0.7 &87.4$\pm$0.5 &88.9$\pm$0.3 &90.0$\pm$0.5 & 90.4$\pm$0.2&\\
\hline Model&\multicolumn{6}{c}{ResNet-18}\\
\hline
Random& 88.5$\pm$0.5 & 88.0$\pm$0.6 & 88.6$\pm$0.4 & 90.1$\pm$0.2 & 90.5$\pm$0.3 & 91.0$\pm$0.3 & \multirow{2}{*}{91.0$\pm$0.3}\\
DsCo& 89.0$\pm$0.4 & 89.4$\pm$0.5 & 89.3$\pm$0.5 & 91.0$\pm$0.2 & 91.2$\pm$0.3 & 91.1$\pm$0.3 & \\
\bottomrule
\end{tabular}
\end{table*}

\begin{table*}
\setlength{\tabcolsep}{9pt}
\caption{Low-IPC Performance comparison on ImageNet-1k. The best performances are denoted in \textbf{bold}.}
\label{tab:img1k_lowipc}
\begin{tabular}{c|cc|cc|cc}
\toprule
\multirow{2}{*}{Method} & \multicolumn{2}{c}{ResNet-18}&\multicolumn{2}{c}{MobileNet-V2}&\multicolumn{2}{c}{EfficientNet-B0}\\
&IPC10&IPC50&IPC10&IPC50&IPC10&IPC50\\
\hline
\multicolumn{7}{c}{Data-Accessible}\\ \hline
RDED&42.0$\pm$0.1&56.5$\pm$0.1&40.4$\pm$0.1&53.3$\pm$0.2&31.0$\pm$0.1&58.5$\pm$0.4\\
DiT-IGD&45.5$\pm$0.5&59.8$\pm$0.3&39.2$\pm$0.2&57.8$\pm$0.2&47.7$\pm$0.1&62.0$\pm$0.1\\
Minimax-IGD&46.2$\pm$0.6&60.3$\pm$0.4&39.7$\pm$0.4&58.5$\pm$0.3&48.5$\pm$0.1&62.7$\pm$0.2\\
DsCo(ours)&\textbf{47.0$\pm$0.2}&\textbf{60.4$\pm$0.1}&\textbf{42.3$\pm$0.2}& \textbf{59.0$\pm$0.2}&\textbf{50.9$\pm$0.1}&\textbf{63.0$\pm$0.1}\\
\hline \multicolumn{7}{c}{Data-Free}\\ \hline
$\mathrm{SRe^2L}$&21.3$\pm$0.6&46.8$\pm$0.2&10.2$\pm$2.6&31.8$\pm$0.3&11.4$\pm$2.5&34.8$\pm$0.4\\
DFDsCo(ours)&43.3$\pm$0.1&58.2$\pm$0.2& 40.6$\pm$0.2& 57.4$\pm$0.1 &49.1$\pm$0.1 & 61.8$\pm$0.2\\
\bottomrule
\end{tabular}
\end{table*}

\begin{figure}
  \centering
  \includegraphics[width=1.0\linewidth]{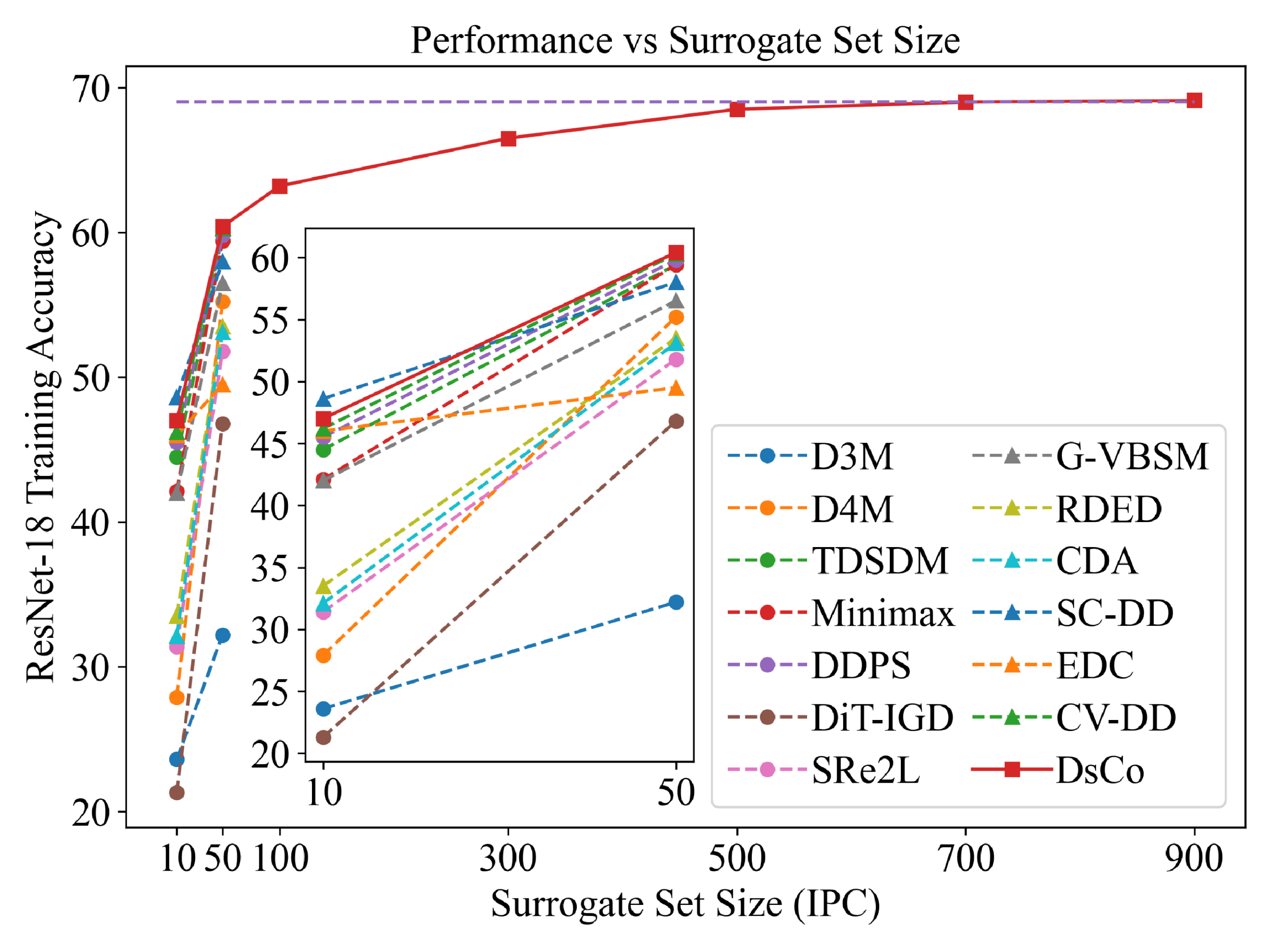}
  \caption{ResNet-18 Accuracy vs IPC for ImageNet-1k. The sub-figure enclosed in the major figure depicts the 10IPC and 50IPC performances, which are commonly reported in the dataset distillation literature.}
  \label{fig:img1k_highipc}
\end{figure}

\subsubsection{Dataset Concentration on ImageNette and ImageWoof} 
Table~\ref{tab:nette_lowipc} and Table~\ref{tab:woof_lowipc} report the performance of the proposed Dataset Concentration (DsCo) method with very small surrogate set sizes (10IPC, 50IPC, and 100IPC) on the ImageNette and ImageWoof datasets, respectively. Following previous literature~\cite{IGD, Minimax, OT}, the accuracies reported in the table are the average evaluation performances of the corresponding architectures trained on the concentrated (or condensed) datasets with one-hot hard labels over five training processes. Under the low-IPC setting, the concentrated dataset is solely synthesized. As illustrated in the table, our concentrated dataset demonstrates strong training performances across both datasets. It achieves SOTA performances in most scenarios, especially for the ResNetAP-10, 10IPC experiment on ImageWoof, surpassing the previous SOTA by $3.8\%$. On ImageNette, it demonstrates comparable performances to previous SOTA, surpassing it by $1.8\%$ for the 10IPC, ResnetAP-10 setting. In particular, it consistently demonstrates superior performance compared to the Vanilla DiT, especially under 100IPC setting, where it demonstrates a $10.5\%$ improvement. The strong performances of the low-IPC concentrated datasets, which are all synthesized via the NOpt method, demonstrate that mitigating the random sampling bias results in significant performance improvement, providing an empirical validation for the effectiveness of the proposed NOpt method.

In Table~\ref{tab:nette_highipc} and Table~\ref{tab:woof_highipc}, we present the dataset concentration performance in high-IPC settings. For the high-IPC concentrated datasets that contain both synthesized samples and selected real samples, we adopt the relabeling technique with ResNet-18 soft labels. Since no previous Dataset Distillation works have reported their high-IPC performances due to the prohibitive generation costs, the high-IPC performance is compared against random sampling (Random). As outlined in the table, the concentrated dataset demonstrates superior performances compared to random sampling, achieving lossless performances at 700IPC for both datasets when evaluated on strong models such as ResNetAP-10 and ResNet-18, where the full dataset training accuracies are within or below the error ranges of the DsCo methods, reaching lossless data concentration rates of $54.3\%$ and $56.2\%$. 

\subsubsection{Dataset Concentration on ImageNet-1k}
Table~\ref{tab:img1k_lowipc} presents the training performances of a series of models (ResNet-18, ResNet-101, MobileNet-V2, and EfficientNet-B0) for the dataset concentration method in the low-IPC setting. The concentrated datasets in the low-IPC setting are all synthesized with the NOpt method. For all ImageNet-1k experiments, we follow the previous literature to use the relabeling technique with ResNet-18 soft labels, and compare the concentrated dataset with the reported values in the corresponding literature of RDED, IGD, and $\mathrm{SRe^2L}$. The table shows that the concentrated dataset achieves SOTA performances among existing dataset distillation methods on all architectures for 10IPC and 50IPC settings. 

To assess the performance of DsCo under extended conditions with a broad range of synthesis IPC conditions, we further plot the ResNet-18 validation accuracy against the dataset size for the concentrated dataset as well as all available reported performances of a variety of baseline methods in Figure~\ref{fig:img1k_highipc}. To the best of our knowledge, no previous works in the dataset distillation literature have reported more than 50IPC performance on ImageNet-1k under the present evaluation setting with ResNet-18 relabeling, potentially due to the prohibitive cost of synthesizing more than 50,000 high-resolution images, nor have lossless performances been reported on ImageNet-1k or its subsets. In contrast, the DsCo method easily scales to high-IPC settings, and it demonstrates lossless performance from 700IPC, achieving a lossless compression rate of $53\%$, almost reducing the challenging ImageNet-1k dataset by half with no performance degradation.

Moreover, in Table~\ref{tab:700generalize}, we report the cross-architecture training performance of the 700-IPC synthetic dataset, whose confusion score is computed with the ResNet-18 architecture. The models are trained with the soft-label relabeling technique with ResNet-18 soft labels. As demonstrated in the table, the 700-IPC concentrated dataset demonstrates the same performance as the full ImageNet-1k dataset on ResNet-18, ResNet-34, and ResNet-101. On EfficientNet-B0 and DenseNet-121, the concentrated dataset demonstrates minor performance degradations of $0.4\%$ on both architectures. In summary, the concentrated dataset demonstrates strong cross-architecture transferability.

\begin{table}
\caption{ImageNet-1k cross-architecture performance of the 700IPC concentrated dataset on ResNet-18, ResNet-34, ResNet-101, EfficientNet-B0, and DenseNet-121.}
\label{tab:700generalize}
\begin{tabular}{c|cc}
\toprule
Model& DsCo Accuracy& Full Accuracy\\ \hline
ResNet-18& 69.0$\pm$0.1&69.0$\pm$0.1\\
ResNet-34& 69.1$\pm$0.1&69.1$\pm$0.2\\
ResNet-101& 70.1$\pm$0.1&70.1$\pm$0.1\\
EfficientNet-B0& 66.4$\pm$0.2&66.8$\pm$0.3\\
DenseNet-121&67.2$\pm$0.2&67.6$\pm$0.2\\
\bottomrule
\end{tabular}
\end{table}

\subsubsection{Performance in Data-Free Scenarios}
In the data-free scenario, the data-free dataset concentration method synthesizes an informative surrogate dataset for the inaccessible target dataset. As the real target data is inaccessible, all the samples are synthesized in this case. Consequently, the generation cost makes it difficult to scale to high-IPC settings. We report the low-IPC performances of the data-free dataset concentration method (denoted as "DFDsCo") on ImageNet-1k and its subsets (ImageNette and ImageWoof) in Table~\ref{tab:img1k_lowipc}, Table~\ref{tab:nette_lowipc}, and Table~\ref{tab:woof_lowipc}, respectively. Remarkably, the data-free dataset concentration method outperforms a series of methods that require data access, generally outperforming Random Sampling, DM, and IDC-I on ImageNette and ImageWoof. Its performances are similar to those of Minimax on the two datasets. On ImageNet-1k, it consistently outperforms RDED on all architectures and both 10IPC and 50IPC settings. Further, among existing data-free methods, the data-free dataset concentration method achieves SOTA performance, significantly surpassing its counterpart ($\mathrm{SRe^2L}$) by $22.0\%$ and $11.4\%$ under 10IPC and 50IPC settings. In summary, the data-free dataset concentration method synthesizes informative surrogate datasets for the target dataset when the target dataset is inaccessible, making a great contribution to information freedom. 

\subsection{Ablation}
\subsubsection{Component Ablation Analysis}
\paragraph{Data-Accessible Low-IPC Ablation}
In the low-IPC setting with data access, the samples are solely synthesized by the NOpt method with data-accessible distribution alignment. In Table~\ref{tab:condenseablation}, we analyze the importance of the two noise-optimization components, namely, the distribution alignment loss and the reality constraint. As illustrated in the table, eliminating either component results in degraded distillation performances. Therefore, both components are vital for the mitigation of the random sampling bias. In particular, the distribution alignment loss makes a great contribution to the performance of the synthesized dataset, as it leads to significant performance improvements for all IPC settings. Meanwhile, the reality constraint provides vital complementary information for the dataset to be both informative and authentic.
\begin{table}
\caption{Component ablation analysis for low-IPC Dataset Concentration}
\label{tab:condenseablation}
\begin{tabular}{cc|ccc}
\toprule
\multicolumn{2}{c}{Components} & \multicolumn{3}{c}{ImageNette, ResNetAP-10}  \\ \hline
$\mathcal{L}_\mathrm{align}$ & $\mathcal{L}_\mathrm{real}$ & Acc@10 & Acc@50 & Acc@100 \\ \hline
$\times$ & $\times$ & 64.8 & 75.5& 80.5\\
$\times$ & $\checkmark$& 64.6& 75.6 & 81.1 \\
$\checkmark$ & $\times$ & 68.9 & 81.4 & 85.5 \\
$\checkmark$ & $\checkmark$ & \textbf{69.8} &\textbf{84.0} & \textbf{86.9} \\ 
\bottomrule
\end{tabular}
\end{table}

\paragraph{Data-Accessible High-IPC Ablation}
Under high-IPC settings, the concentrated dataset is a combination of the samples synthesized with the Noise-Optimization (NOpt) method and the samples selected from the target dataset via Doping. To analyze the contributions of the two stages, we conduct a component ablation analysis on ImageNette, where we independently eliminate the two stages and replace the corresponding eliminated samples with randomly selected samples from the original dataset. When the NOpt samples are replaced by random samples, but Doping persists, we recompute the confusion scores using the random samples for the subsequent Doping. The ResNet-18 evaluation accuracies are outlined in Table~\ref{tab:stageablation}. As indicated by the table, eliminating either stage results in globally degraded performances across all IPC settings. Therefore, both stages play a vital role in the Dataset Concentration method. Notably, the two stages contribute jointly to the performance of the concentrated dataset, as the performance degradations caused by the elimination of either stage are similar.
\begin{table}
\caption{Stage ablation analysis for high-IPC Dataset Concentration}
\label{tab:stageablation}
\begin{tabular}{cc|cccc}
\toprule
\multicolumn{2}{c}{Components} & \multicolumn{4}{c}{ImageNette, ResNet-18}  \\ \hline
NOpt & Doping & 200 & 300 & 400 & 700 \\ \hline
$\times$ & $\times$ & 95.1& 95.6 & 96.1 & 97.0 \\
$\times$ & $\checkmark$& 95.5& 95.4&96.4 & 97.3\\
$\checkmark$ & $\times$ &95.4& 95.5 & 96.4& 97.2\\
$\checkmark$ & $\checkmark$ & \textbf{96.0} &\textbf{96.1} & \textbf{96.9} & \textbf{97.9} \\ 
\bottomrule
\end{tabular}
\end{table}

\paragraph{Data-Free Ablation}
When the target dataset is inaccessible, we can synthesize a small and informative surrogate dataset with the data-free variant of noise-optimization. To evaluate the effect of each component in the data-free dataset concentration, we independently eliminate the three losses, $\mathcal{L}_\mathrm{stats}$,  $\mathcal{L}_\mathrm{maxoc}$, and $\mathcal{L}_\mathrm{real}$ (with $t$ dropped for clarity),  and we evaluate the generated surrogate datasets of ImageNette with ResNetAP-10. Specifically, the model is trained on the corresponding datasets with the SGD (Bottou, 2018~\cite{SGD}) optimizer. Table \ref{tab:dfdcabl} shows that removing any components degrades the performance, especially under 10IPC. The results indicate that optimal data-free dataset concentration performance requires the synergistic use of all three constraints.
\begin{table}
\setlength{\tabcolsep}{5pt}
\caption{Component ablation analysis of Data-Free Dataset Concentration. In this experiment, the evaluation model is trained with the SGD~\cite{SGD} optimizer.}
\label{tab:dfdcabl}
\begin{tabular}{ccc|cc}
\toprule
\multicolumn{3}{c}{Components} & \multicolumn{2}{c}{ImageNette, ResNetAP-10}  \\ \hline
$\mathcal{L}_\mathrm{stats}$ & $\mathcal{L}_\mathrm{maxoc}$ & $\mathcal{L}_\mathrm{real}$ & Acc@10IPC & Acc@50IPC \\ \hline
$\checkmark$ & $\times$ & $\times$ & 60.3& 76.2\\
$\times$ & $\checkmark$ & $\times$ & 61.7& 74.1 \\
$\times$ & $\times$ & $\checkmark$ & 58.5&74.5 \\
$\times$ & $\checkmark$ & $\checkmark$ & 61.5&74.9  \\
$\checkmark$ & $\times$ & $\checkmark$ &60.9 &74.6 \\
$\checkmark$ & $\checkmark$ & $\times$ &61.9 & 76.2 \\ 
$\checkmark$ & $\checkmark$ & $\checkmark$ & \textbf{65.0}&\textbf{78.7}\\ 
\bottomrule
\end{tabular}
\end{table}

\subsubsection{Robustness Analysis}
\paragraph{Data-Accessible Dataset Concentration}
The data-accessible dataset concentration framework has been established with great robustness, such that there is only one hyperparameter that requires adjustment. That is, the relative strength of the distribution alignment loss compared to the reality constraint, $\lambda_\mathrm{align}$. The performance of the dataset concentration framework under low-IPC setting with varying hyperparameter $\lambda_\mathrm{align}$ is listed in Table~\ref{tab:hpalign}. As outlined in the table, the low-IPC performance is robust against variation in $\lambda_\mathrm{align}$ on a logarithmic scale, with a minor performance degradation of $1\%$ on average when dividing $\lambda_\mathrm{align}$ by 10 from 5e-4 to 5e-5. 
In this work, the same optimal $\lambda_\mathrm{align}$ works for all datasets and all IPC settings. For high-IPC settings, the condensed dataset is synthesized with the optimal $\lambda_\mathrm{align}$ and subsequently participates in the doping process with no further hyperparameter adjustments.
\begin{table}
\caption{Hyperparameter sensitivity analysis for low-IPC Dataset Concentration}
\label{tab:hpalign}
\begin{tabular}{c|ccc}
\toprule
\multirow{2}{*}{$\lambda_\mathrm{align}$}& \multicolumn{3}{c}{ImageNette, ResNetAP-10}  \\ 
& Acc@10 & Acc@50 & Acc@100 \\ \hline
5e-2& 58.6 & 80.8& 80.6\\
5e-3 & 66.6 & 81.6& 86.2 \\
5e-4 & \textbf{69.8} & \textbf{84.0} & \textbf{86.9} \\
5e-5 & 69.0 &83.0 & 85.8 \\ 
5e-6 & 67.8 &78.4 & 82.2 \\ 
\bottomrule
\end{tabular}
\end{table}

\paragraph{Data-Free Dataset Concentration}
In the data-free dataset concentration framework, as indicated by Equation~\ref{eqn:dfalign} and Equation~\ref{eqn:nopt}, the balance between $\mathcal{L}_\mathrm{stats}$, $\mathcal{L}_\mathrm{maxoc}$, and $\mathcal{L}_\mathrm{real}$ determines the concentrated dataset that is solely generated with the Noise-Optimization process. As previously argued, $\mathcal{L}_\mathrm{real}$ has a fixed weight of 1.0 as optimization is scale-invariant to the sum of losses. The weight of $\mathcal{L}_\mathrm{real}$ is fixed to the maximal value that enables the synthesis of visually recognizable samples in the absence of $\mathcal{L}_\mathrm{stats}$, which can be easily determined with a visual assessment, and is fixed to 10.0 for all experiments in this work. Therefore, only $\mathcal{L}_\mathrm{stats}$ requires empirical adjustment. The performances of the data-free-concentrated dataset under different strengths of $\mathcal{L}_\mathrm{stats}$ are listed in Table~\ref{tab:hpstats}. As indicated by the table, the data-free dataset concentration method is robust to hyperparameter variation on a logarithmic scale, as multiplying $\lambda_\mathrm{stats}$ by 10 only degrades the performance by $0.5\%$.
\begin{table}
\caption{Hyperparameter sensitivity analysis for low-IPC Dataset Concentration in data-free scenario (DFDsCo). In this experiment, the models are trained with the SGD~\cite{SGD} optimizer.}
\label{tab:hpstats}
\begin{tabular}{c|cc}
\toprule
\multirow{2}{*}{$\lambda_\mathrm{stats}$}& \multicolumn{2}{c}{ImageNette, ResNetAP-10}  \\ 
& Acc@10 & Acc@50 \\ \hline
1.0& 58.4& 72.0\\
0.1& 61.0 & 73.9\\
0.01 & 64.6 & 78.2 \\
1e-3 & \textbf{65.0} & \textbf{78.7} \\
1e-4 & 63.8 &76.1 \\ 
1e-5 & 60.9 &75.0 \\ 
\bottomrule
\end{tabular}
\end{table}

\subsubsection{Cost Analysis}
To evaluate the cost of composing the concentrated dataset, we separately perform cost analyses in the low-IPC and high-IPC settings. 
\paragraph{Low-IPC Synthesis Cost}
Under low-IPC setting, all the samples are synthesized. We present the total synthesis cost (including preparation cost and generation cost) of the proposed framework under data-accessible and data-free scenarios (DsCo and DFDsCo) in Table~\ref{tab:costs_lowipc}, with recent open-source reproducible diffusion-based methods (Minimax, IGD-DiT) as baselines. The costs are measured in the maximal memory requirement and the running time on NVIDIA RTX-2080Ti GPUs to synthesize 10IPC, 50IPC, and 100IPC datasets for ImageNet-1k. For the Minimax, IGD-DiT, DsCo, and DFDsCo methods, the preparation stages are fine-tuning DiT, training the classifier, encoding samples, and generating template samples, respectively. Note that the GPU running time for IPC100 of DsCo is for illustration only, as the Doping has been triggered at 100IPC for ImageNet-1k. 

As illustrated by the table, among all reproducible diffusion-based methods, DsCo and DFDsCo demonstrate high synthesis efficiency under low-IPC settings, with the lowest total synthesis times and peak memory requirements. Their low memory requirements allow them to be run on a single NVIDIA GeForce RTX2080 Ti GPU, while the Minimax and DiT-IGD methods require at least 2 such GPUs.
\begin{table*}
\caption{Low-IPC synthesis cost analysis of ImageNet-1k on NVIDIA GeForce RTX2080 Ti GPUs.}
\label{tab:costs_lowipc}
\begin{tabular}{c|c|ccc|ccc|c}
\toprule
\multirow{2}{*}{Method}& \multirow{2}{*}{Preparation} & \multicolumn{3}{c}{Generation} & \multicolumn{3}{c}{Total} & \multirow{2}{*}{Peak Memory} \\ 
& & IPC10 & IPC50 & IPC100 & IPC10 & IPC50 & IPC100 & \\ \hline
Minimax & 253h & 7h & 35h & 70h & 260h & 288h & 323h & 14.8G \\
DiT-IGD & 124h & 69h & 347h & 694h & 193h & 471h & 818h & 13.2G \\
DsCo & 14h & 28h & 62h & 119h & 42h & 76h & 133h & 4.3G \\
DFDsCo & 142h & 15h & 53h & 106h & 157h & 195h & 247h & 4.7G \\
\bottomrule
\end{tabular}
\end{table*}

\begin{figure}
  \centering
  \includegraphics[width=1.0\linewidth]{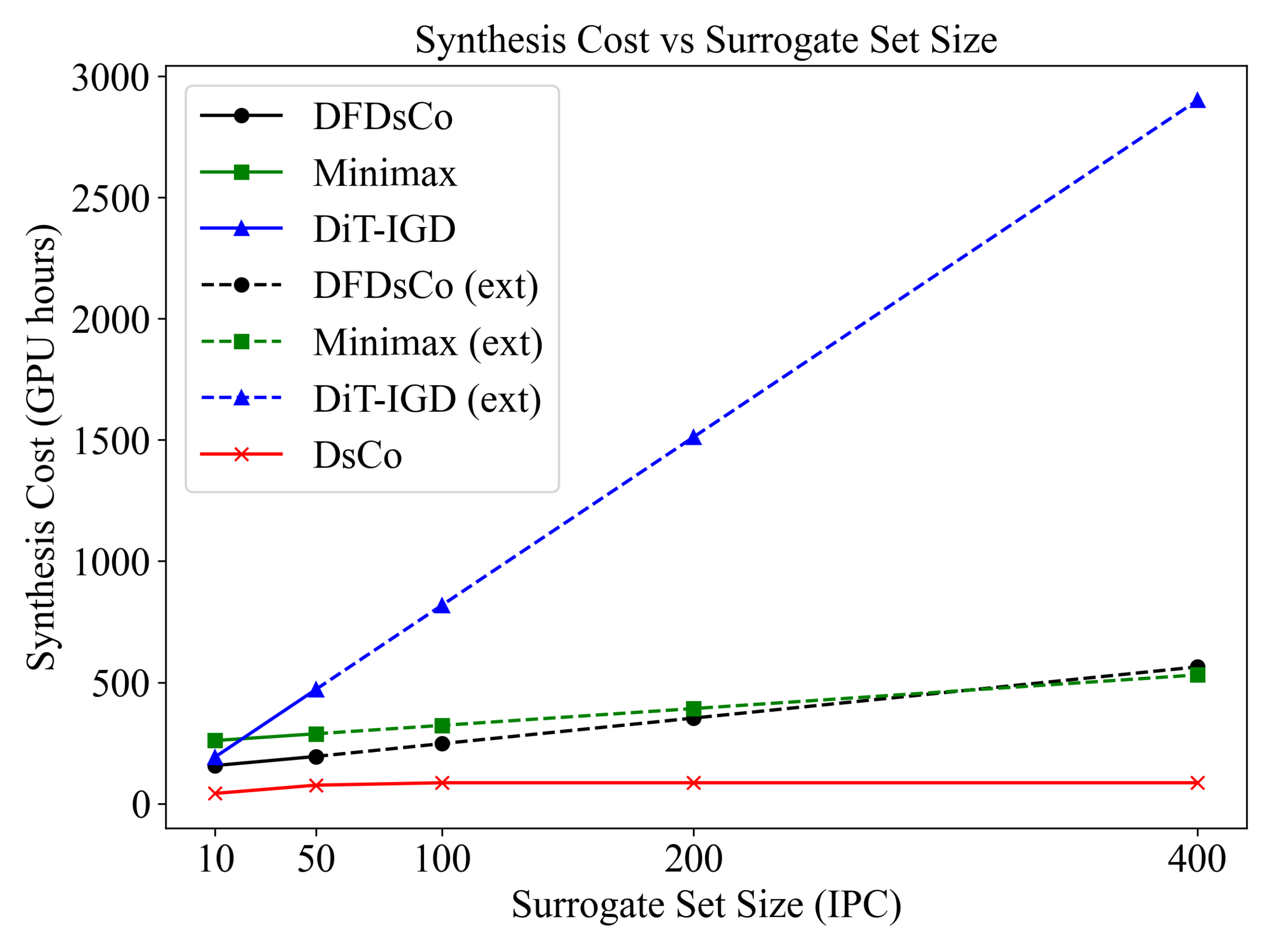}
  \caption{Synthesis costs, measured in running time on NVIDIA GeForce RTX 2080 Ti GPUs, plotted against IPC for contemporary open-source diffusion-based dataset distillation or concentration methods. The dashed lines are estimated costs obtained through linear extrapolation.}
  \label{fig:cost_highipc}
\end{figure}

\paragraph{High-IPC Synthesis Cost}
Figure~\ref{fig:cost_highipc} plots the total synthesis cost for the Minimax, DiT-IGD, DsCo, and DFDsCo methods for extended IPC settings. In particular, the costs for Minimax, DiT-IGD, and DFDsCo beyond 100IPC are anticipated values obtained through linear extrapolation, thus expressed in dashed lines. The extrapolation works because their synthesis costs scale linearly to the number of samples. In contrast, as illustrated in the figure, the synthesis cost of DsCo ceases to increase after 100IPC, where the doping has been triggered. This is because the cost of Doping is fixed to the constant cost of training the model on the synthesized dataset, labeling the target dataset with the model, computing the confusion score, and sorting the samples according to their confusion scores. It does not depend on the number of samples to be selected. Therefore, the DsCo framework enjoys superior extensibility over other diffusion-based dataset distillation methods.

\subsubsection{Visualization}
\begin{figure}
  \centering
  \includegraphics[width=1.0\linewidth]{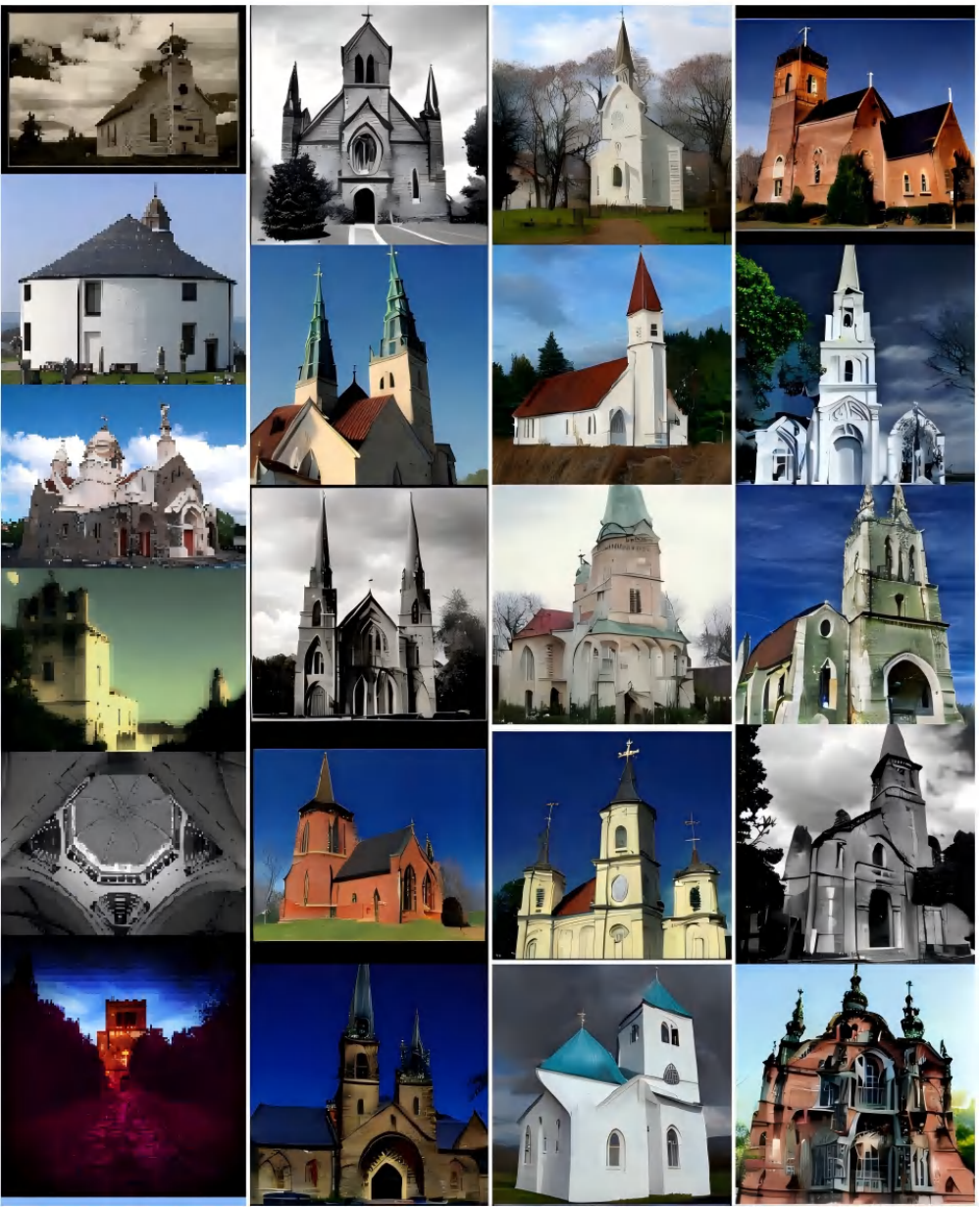}
  \caption{Visualization of the real samples (left), DiT synthetic samples (middle left), DsCo synthetic samples (middle right), and the DFDsCo synthetic samples (right) of the Church class of the ImageNette dataset.}
  \label{fig:church}
\end{figure}

\begin{figure}
  \centering
  \includegraphics[width=1.0\linewidth]{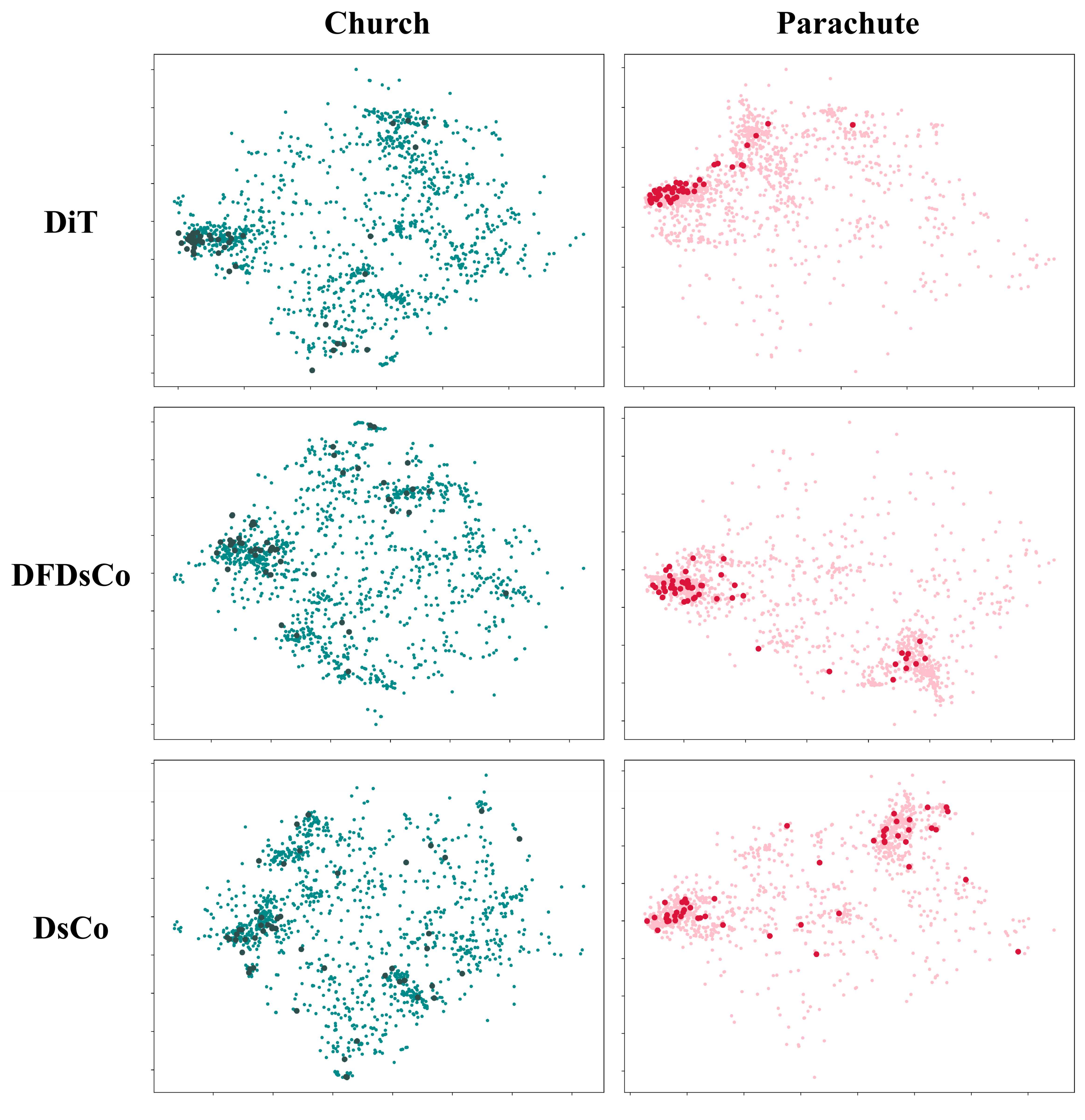}
  \caption{The t-SNE distribution visualization for the 'church' (left) and 'parachute' (right) classes of ImageNette. The latent codes of the synthetic samples are compressed to 2-D and plotted on top of the compressed latent codes of real samples. The synthetic samples are synthesized with DiT (top), DFDsCo (middle), and DsCo (bottom).}
  \label{fig:distribution_lowipc}
\end{figure}

\begin{figure*}
  \centering
  \includegraphics[width=1.0\linewidth]{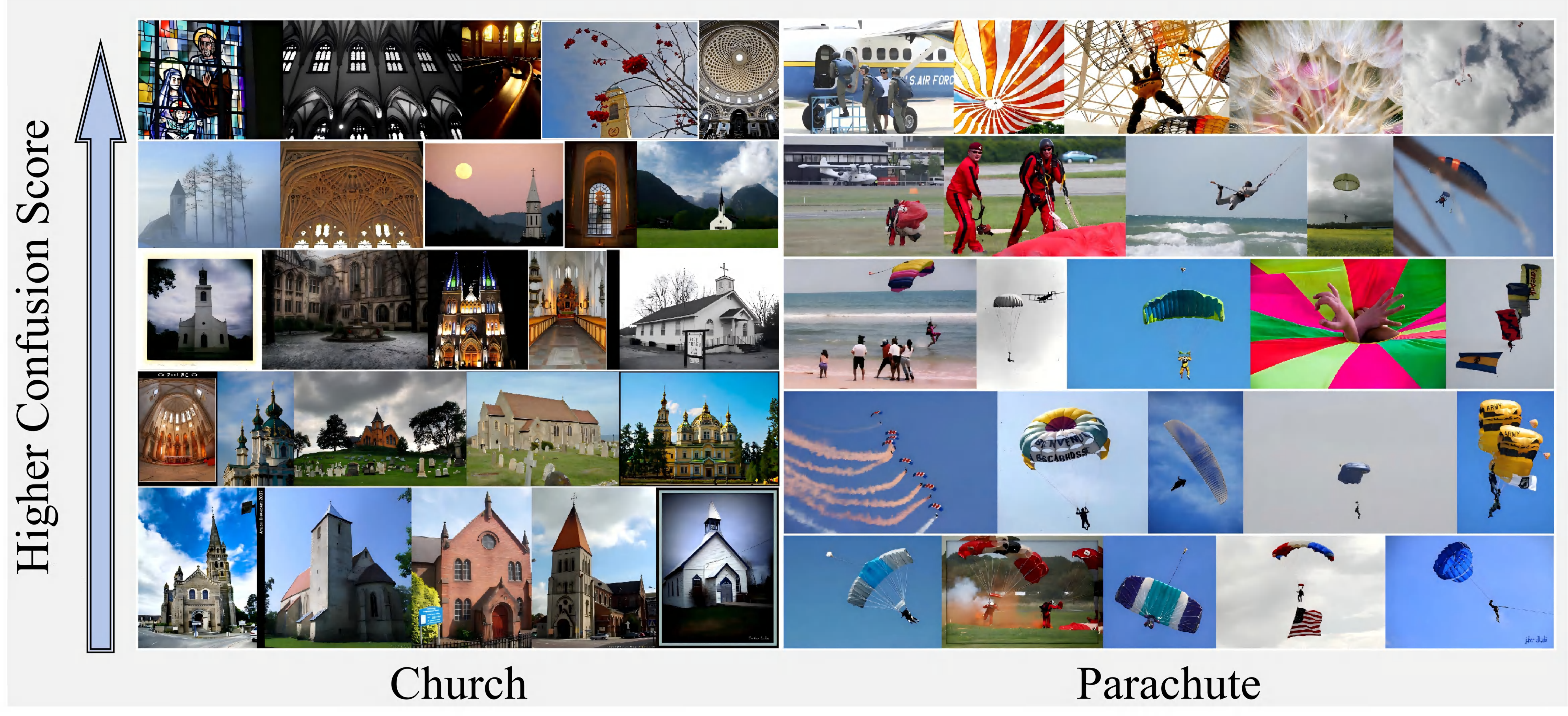}
  \caption{The real samples with increasing confusion scores for the 'church' and 'parachute' classes of the ImageNette dataset.}
  \label{fig:samples_highipc}
\end{figure*}

\begin{figure}
  \centering
  \includegraphics[width=1.0\linewidth]{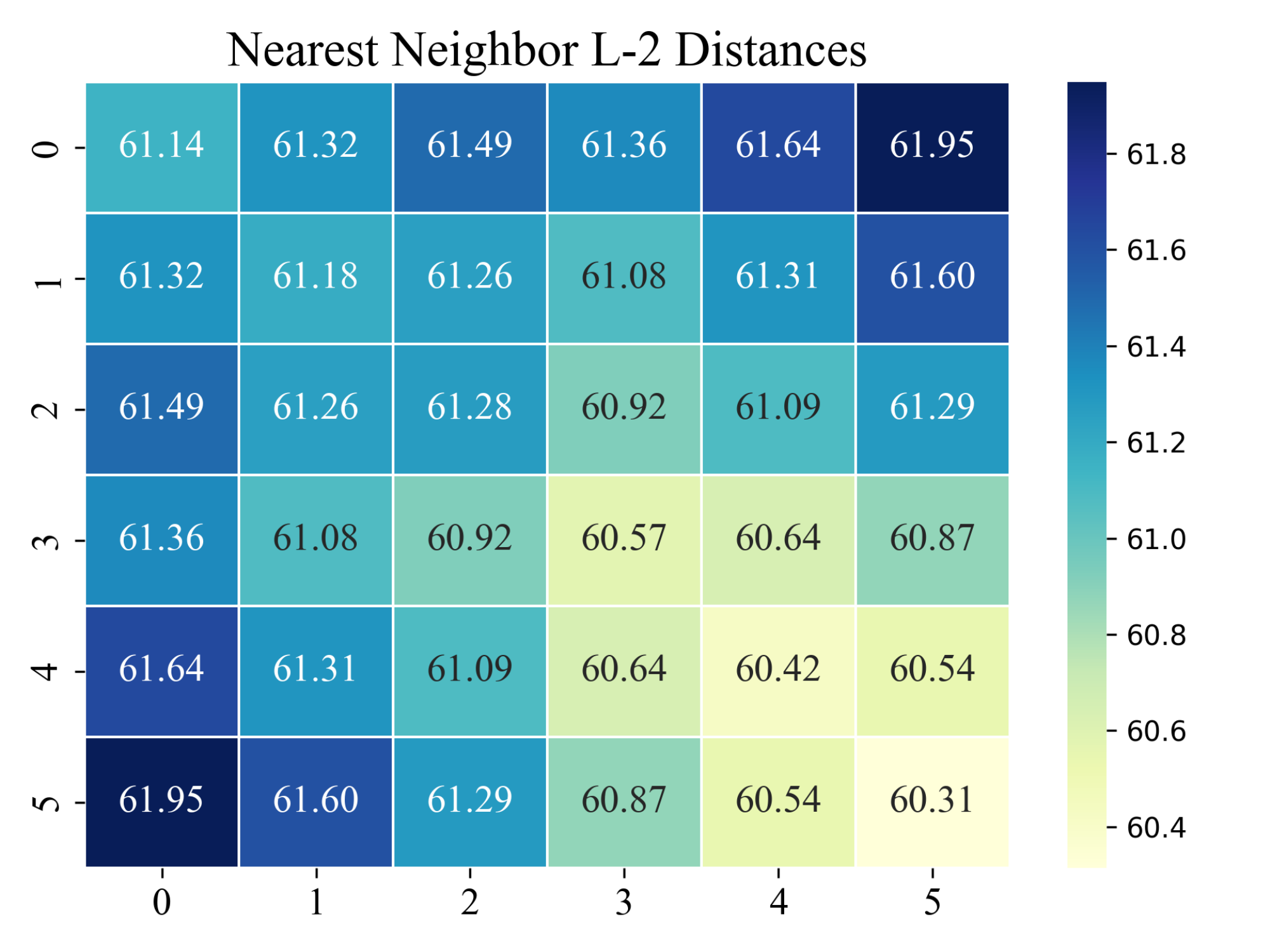}
  \caption{The average intra-class latent-space nearest-neighbor mutual L-2 distance for five groups of target samples of descending confusion scores computed within each group (diagonal) and between different groups (off-diagonal).}
  \label{fig:mutl2_heatmap}
\end{figure}

\paragraph{Low-IPC Visualization}
Under low-IPC settings, the Doping is not triggered, and all the samples are synthesized using the corresponding Noise-Optimization methods with or without data access. Presented in Figure~\ref{fig:church} is the visualization of random real samples (left), DiT-generated samples (middle left), the DsCo samples (middle right), and the data-free concentrated samples (right) in the low-IPC setting of the 'Church' class of ImageNette. As illustrated by the images, all samples enjoy great authenticity, demonstrating the benefit of synthesizing images with pre-trained diffusion models. 

Figure~\ref{fig:distribution_lowipc} depicts the t-SNE (Laurens, 2008~\cite{TSNE}) low-dimensional sample distribution of the synthesized latent codes plotted on top of that of the VAE-encoded latent codes of the target dataset of the 'church' and 'parachute' classes of ImageNette. For visual comparison, the vanilla DiT latent codes are compared against DFDsCo and DsCo latent codes. As illustrated in the figure, the DiT samples tend to concentrate in dense regions of the target dataset distribution, as observed in previous literature (Gu, 2024~\cite{Minimax}). It concentrates too much on the densest area, and thus cannot cover other dense regions of lower concentration. Both DFDsCo and DsCo demonstrate improved coverage, with their synthesized latent codes scattered in multiple dense areas. Since DFDsCo does not access the target dataset, its capability to find the dense regions indicates that the correct distribution has been memorized by the diffusion model, and that the proposed method is capable in uncovering this hidden memory. Among the three methods, DsCo generally performs better than its two counterparts, occupying more clustered regions than DFDsCo and DiT, especially for the 'parachute' class.

\paragraph{Visualization for high-IPC settings}
A visualization analysis is performed for the samples sampled with the doping method under high IPC settings. Figure~\ref{fig:samples_highipc} displays the target samples with increasing confusion scores of the 'church' and 'parachute' classes of ImageNette. As illustrated by the figure, the visual dissimilarity between the target samples in the same group increases as the confusion score increases, and the high-confusion samples are very dissimilar from the low-confusion samples. This substantiates the previous argument that a high confusion score is associated with high dissimilarity of the sample to its neighbors, which indicates a high chance of being a far-apart sample. 

Further, the average intra-class latent-space mutual L-2 distances of the groups of samples with increasing confusion scores for the ImageNette dataset are visualized in the heatmap in Figure~\ref{fig:mutl2_heatmap}. The samples in ImageNette are sorted by their confusion scores in descending order, and the first $1000\times 10$ samples are evenly separated into 5 groups, $(1,2,3,4,5)$, with descending confusion scores. The heatmap presents the average intra-class latent-space mutual L-2 distances of all combinations of the five groups. The latent codes are encoded with the VAE from stable-diffusion (Robin, 2021~\cite{sd}). As indicated by the figure, there is a significant leap in the average mutual L-2 distance within the same group of samples (i.e., the diagonal values) between groups $(0,1,2)$ to groups $(3,4,5)$, which substantiates the assumption that samples with high confusion scores are likely to be far-apart samples. Further, the average mutual distances between group 0 and groups 1,2,3,4,5 monotonically increases (the combination is denoted as $(0;1,2,3,4,5)$, where the groups after the semicolon are listed in ascending order of average latent-space L-2 distances from samples in group 0.), and the same tendency goes for the group combinations $(1;3,2,4,5)$, $(2;3,4,5)$, $(3;4,5,6)$, and $(4;5,6)$. This indicates that the proposed confusion score is positively associated with the degree to which a sample is far away from the majority of samples. Therefore, the confusion score serves as a good indicator of the degree to which a sample is both far-away and far-apart from other samples.

\section{Conclusion}
In this work, we have addressed three critical limitations in dataset distillation: the lack of a theoretical foundation, the inefficiency in high-IPC settings, and the inability to operate without access to the original data. Firstly, we have established a theoretical framework to analyze the dataset distillation problem. The theoretical analysis demonstrates that dataset distillation is equivalent to a distribution matching problem, justifying the use of diffusion models in the dataset distillation task. Through further analysis, we have identified a random bias impairing the training performance of diffusion-synthesized samples, and revealed a fundamental efficiency limit inherent to the dataset distillation paradigm stemming from the far-apart samples in target datasets, providing a theoretical explanation for the difficulty in scaling dataset distillation methods to large data volumes. In light of the analysis, we proposed the Dataset Concentration (DsCo) framework, which adopts a diffusion-based Noise-Optimization (NOpt) method applicable under both data-accessible and data-free scenarios to synthesize informative samples through the denoising process with mitigated random sampling biases. DsCo further incorporates an optional "Doping" process for high data volumes, which samples the far-apart samples in the target dataset to overcome the efficiency limitation of data synthesis. Extensive experiments demonstrate that DsCo achieves state-of-the-art performance under multiple settings on various datasets. Crucially, it nearly reduces the dataset size by half with no performance degradation, proving its superior scalability. Furthermore, in the challenging data-free setting, DsCo outperforms all existing methods, offering a practical solution for privacy-sensitive applications. In summary, this work advances dataset distillation by providing a solid theoretical ground, an efficient and adaptable concentration framework, and substantial empirical validation, paving the way for more trustworthy and practical data-efficient learning.

\backmatter

\bmhead{Supplementary information}

If your article has accompanying supplementary file/s please state so here. 

Authors reporting data from electrophoretic gels and blots should supply the full unprocessed scans for key as part of their Supplementary information. This may be requested by the editorial team/s if it is missing.

Please refer to Journal-level guidance for any specific requirements.




\section*{Declarations}

\begin{itemize}
\item \textbf{Funding} This work was supported by Beijing Natural Science Foundation (JQ24022), the National Natural Science Foundation of China (No. 62192785, No. 62372451, No. 62372082, No. 62272125, No. 62306312, No. 62036011, No. 62192782), CAAI-Ant Group Research Fund (CAAI-MYJJ 2024-02), Young Elite Scientists Sponsorship Program by CAST (2024QNRC001), the Project of Beijing Science and technology Committee (Project No. Z231100005923046).
\item \textbf{Competing interests} The authors have no competing interests to declare that are relevant to the content of this article.
\item \textbf{Ethics approval and consent to participate} Not Applicable.
\item \textbf{Consent for publication} All authors have approved the submission of this manuscript for publication.
\item \textbf{Data availability} The datasets used to synthesize the concentrated datasets in this study are publicly available, as cited in the paper. The concentrated datasets for ImageNet-1k under 10IPC and 50IPC settings in data-accessible and data-free scenarios are available at the following repository: https://pan.baidu.com/s/1WyR47H-cG06Zm3WfLpFm7Q with the access code 1234, and the complementary real samples can be sampled from the publicly available dataset using the code provided below. The authors are willing to provide any additional data upon request.
\item \textbf{Materials availability} Not Applicable.
\item \textbf{Code availability} The implementation code for this study is available at https://github.com/kkkkqq/Dataset-Concentration
\item \textbf{Author contribution} All authors contributed to the study conception and design. Material preparation, data collection and analysis were performed by Tongfei Liu and Yufan Liu. The first draft of the manuscript was written by Tongfei Liu, and all authors commented on previous versions of the manuscript. All authors read and approved the final manuscript.
\end{itemize}







\begin{appendices}

\section{Applicability of the Theoretical Framework}\label{app:extendDM}
In Section~\ref{sec:DDmm}, we theoretically proved Proposition~\ref{prop:1}, which states that dataset distillation is a distribution matching task under Assumption~\ref{assum1}. For a particular tuple of $(\Phi, \mathcal{T}, \mathcal{S})$, Assumption~\ref{assum1} assumes that there exists a shift-invariant positive-definite kernel function $k(x^{\tau}, x^s)$ that monotonically increases with the chance of recognizing $x^{\tau}$ by memorizing $x^s$. Under this formulation, it is unclear as to how likely this assumption can be met. In this section, we develop a generalized proposition associated with Proposition~\ref{prop:1} under a looser assumption. 

\subsubsection{Extended Proposition}
For the tuple of $(\Phi, \mathcal{T}, \mathcal{S})$, consider an arbitrary invertible transformation $\chi(\cdot)$, which maps $x^{\tau}\in\mathcal{T}$ and $x^s\in\mathcal{S}$ into $\chi(x^{\tau})$ and $\chi(x^s)$. With a slight abuse of notation, we denote the projected samples as $\chi^{\tau}$ and $\chi^s$, respectively. If there exists a shift-invariant positive-definite kernel function $k_\chi\left(\chi^{\tau}, \chi^s\right)$ in this projected space, which monotonically increases with the chance of recognizing $x^{\tau}$ by memorizing $x^s$, then we can immediately prove that the expected chance of recognition is maximized when $P_\mathcal{S}^\chi = P_\mathcal{T}^\chi$ in the space projected by $\chi$ by simply replacing the $x$ by $\chi(x)$ in the analysis presented in Section~\ref{sec:DDmm}. Since $p^\chi\left(\chi\right)d\chi = p(x)dx$, and the reversibility of $\chi(\cdot)$ ensures that $\frac{|d\chi|}{|dx|}\neq 0$, we then deduce that \begin{equation}
    p_\mathcal{T}^\chi(\chi) = p_\mathcal{S}^\chi(\chi) ~\forall \chi \iff p_\mathcal{T}(x) = p_\mathcal{S}(x) ~\forall x.
\end{equation}
Therefore, we have proved Proposition~\ref{prop:2}. 

\section{Monte-Carlo Illustration of Random Bias}\label{app:noiseMonteCarlo}
This section presents a simplified Monte-Carlo experiment to illustrate the effect of randomness. As argued in Section~\ref{sec:randomsamplebias}, for a high-dimensional Gaussian distribution, the norms of the random Gaussian noises are very close to the square root of their dimension. Therefore, we can ignore the variation in the norms of the noises and only consider the angular distribution. In this case, since the spherical surface is measurable, the surface can be divided into regions of equal area. The spherical homogeneity of the Gaussian distribution then dictates that a random Gaussian noise tensor has an equal chance to fall into any region. Therefore, we can reformulate this Gaussian noise sampling into a simple arrangement problem. For a set of $N$ random Gaussian noise tensors, with the spherical surface divided into $N$ equally-sized regions, the ideal sampling is when each region has one sample, in which case the sample distribution of noise tensors closely aligns with the continuous theoretical distribution. However, the chance of achieving this ideal sampling can be analytically computed to be $\frac{N!}{N^N}$, which monotonically decreases with $N$. 

In practice, we can estimate the average number of regions occupied by at least one random noise by a simple Monte-Carlo experiment using the PyTorch package (Paszke, 2019) through the following command:
\begin{verbatim}
import torch
def average_occupation_rate(
    n_exp, n_smps
    ):
    rands=torch.randint(
        n_smps, 
        (n_exp, n_smps)
        )[:,None,:]
    bins=torch.arange(
        n_smps
        )[None,:,None]
    noempty=(rands==bins).sum(2)>0
    return noempty.sum(1).to(
        torch.float).mean()
\end{verbatim}
The mean and standard deviations of the number of occupied regions is listed in Table~\ref{tab:app_rand} with $\mathrm{n_{exp}}=500$. As illustrated in the table, no less than $35\%$ of the regions are unoccupied, demonstrating a significant distribution deviation resulting from random sampling bias.
\begin{table}
\caption{The means and standard deviations of the number of occupied regions.}
\label{tab:app_rand}
\begin{tabular}{c|ccc}
\toprule
$N$&10&50&100\\
\hline mean& 6.5 & 31.7& 63.4\\
std & 1.0 & 2.2& 3.2\\
\bottomrule
\end{tabular}
\end{table}

\section{Feature Projector}\label{app:projector}
The feature projector used in this work adopts an encoder-like architecture that is built upon the "vanilla VAE" architecture in the publicly available implementation of VAE from Subramanian (2020~\cite{Subramanian2020}). We made a few key modifications to adapt this small encoder into a feature projector. Firstly, we removed the final linear layer as it breaks the spatial shift-invariance by assigning different weights to different spatial locations. Secondly, we removed all the batch normalization layers to avoid mutual interference of features. Thirdly, the hidden dimensions of the encoder layers are set to (256, 512, 1024, 2048), and the first convolutional layer has a smaller stride of 1. Finally, the fully connected convolutional layers are replaced by the grouped convolutional layers with 16 independent groups to reduce the computational cost in feature projection. 

\section{Evaluation Implementations}\label{app:implementation}
In this section, we provide the implementation details of the evaluation methods. 

\subsection{Hard-Label low-IPC Training on ImageWoof and ImageNette}\label{app:hardsetting}
In Table~\ref{tab:nette_lowipc} and Table~\ref{tab:woof_lowipc}, the 10IPC, 50IPC, and 100IPC synthetic samples are used to train a series of models (Convnet6, ResNet-AP10, ResNet-18) for 2000, 1500, and 1000 epochs, respectively. The models are trained with the concentrated or synthesized datasets with one-hot hard labels using an AdamW optimizer with learning rate 0.001, betas (0.9, 0.999), epsilon 1e-8, and weight decay 0.01. A StepLR scheduler with two milestones at $2/3$ and $5/6$ of the total training epochs is applied with gamma=0.2. During training, a RandomResizedCrop (Lee, 2022~\cite{LEE2022143}) of scale (0.5,1.0), a RandomHorizontalFlip of chance 0.5, a ColorJitter of $(0.4,0.4,0.4)$, and a Lightning augmentation (Cubuk, 2019~\cite{cubuk2019autoaugment}) are applied on the dataset with a subsequent cutmix augmentation (Yun, 2019~\cite{yun2019cutmix}) applied at a chance of 1.0, with its beta value set to 1.0 as well. Meanwhile, the batch size of training is universally set to 64. 

\subsection{High-IPC training with relabeling technique on ImageWoof and ImageNette}\label{app:netterelabel}
For the high-IPC setting, we adopt the relabeling technique with the soft labels provided by the official pre-trained ResNet-18 model from the TorchVision (2016~\cite{torchvision2016}). The soft-label training temperature is set to $20.0$, and the total number of epochs is 1000 and 500 for datasets of no more than 200IPC and those with more data, respectively. In high-IPC settings, the same augmentation and optimizer as in Appendix~\ref{app:hardsetting} is applied on the datasets, and it uses a default CosineAnnealingLR scheduler from PyTorch with $T_\mathrm{max}$ equal to twice the total number of epochs. 

\subsection{ImageNet-1k training with relabeling technique}
The ImageNet-1k training setting for low-IPC is the same as the setting in Appendix~\ref{app:netterelabel}, but the ColorJitter and Lightning are removed from the training, and the training epochs is fixed to 300 for both 10IPC and 50IPC. Moreover, the adamW learning rate increases to 0.002 and 0.0025 upon training the EfficientNet-B0 and MobileNet-V2.

The high-IPC training on ImageNet-1k trains the models by 100 epochs, using a different scheduler that linearly increases the learning rate in the first ten epochs of training, starting from 0.1, and subsequently decreases the learning rate to 0 in 90 steps using CosineAnnealingLR. Other settings are the same as the low-IPC setting.




\end{appendices}


\bibliography{bib}

\end{document}